\documentclass[lettersize,journal]{IEEEtran}
\usepackage{color,soul} 
\soulregister\cite7 
\soulregister\citep7 
\soulregister\citet7 
\soulregister\ref7 
\soulregister\pageref7 
\usepackage{amsmath,amsfonts}
\usepackage{array}
\usepackage{booktabs}
\usepackage[caption=false,font=normalsize,labelfont=sf,textfont=sf]{subfig}
\usepackage{textcomp}
\usepackage{stfloats}
\usepackage{verbatim}
\usepackage{graphicx}
\usepackage{cite}
\usepackage{multirow}
\usepackage[pagewise]{lineno}
\usepackage[table,xcdraw]{xcolor}
\usepackage[numbers]{natbib}
\usepackage{float}
\usepackage[colorlinks=true,
            citecolor=blue,
            linkcolor=blue,
            anchorcolor=blue,
            urlcolor=blue]{hyperref}
\usepackage[linesnumbered,ruled]{algorithm2e}
\usepackage{amsthm}

\usepackage{pifont}

\newtheorem{remark}{Remark}

\begin{document}


\title{LifelongPR: Lifelong point cloud place recognition based on sample replay and prompt learning}

\author{
\IEEEauthorblockN{
Xianghong Zou,
Jianping Li, \textit{Member}, \textit{IEEE},
Zhe Chen,
Zhen Cao,
Zhen Dong, \textit{Member}, \textit{IEEE},
Qiegen Liu, \textit{Senior Member}, \textit{IEEE},
Bisheng Yang}

\thanks{
This study was supported by the National Natural Science Foundation Project (No. 42130105, No. 42201477, No. 42171431). (Corresponding author: Jianping Li and Qiegen Liu)

Xianghong Zou is with the School of Advanced Manufacturing, Nanchang University, Nanchang 330031, China, and also with the State Key Laboratory of Information Engineering in Surveying, Mapping and Remote Sensing, Wuhan University, Wuhan 430079, China. (e-mail: ericxhzou@ncu.edu.cn)

Jianping Li is with the School of Electrical and Electronic Engineering, Nanyang Technological University, Singapore 639798. (e-mail: jianping.li@ntu.edu.sg)

Zhe Chen, Zhen Cao, Zhen Dong, and Bisheng Yang are with the State Key Laboratory of Information Engineering in Surveying, Mapping and Remote Sensing, Wuhan University, Wuhan 430079, China. (e-mail: bshyang@whu.edu.cn)

Qiegen Liu is with the School of Information Engineering, Nanchang University, Nanchang 330031, China. (e-mail: liuqiegen@ncu.edu.cn)
}
}

\markboth{Journal of \LaTeX\ Class Files,~Vol.~xxx, No.~xxx, xxx~xxx}%
{Shell \MakeLowercase{\textit{et al.}}:LifelongPR: Lifelong point cloud place recognition based on sample replay and prompt learning}

\maketitle

\begin{abstract}
Point cloud place recognition (PCPR) determines the geo-location within a prebuilt map and plays a crucial role in geoscience and robotics applications such as autonomous driving, intelligent transportation, and augmented reality. In real-world large-scale deployments of a geographic positioning system, PCPR models must continuously acquire, update, and accumulate knowledge to adapt to diverse and dynamic environments, i.e., the ability known as continual learning (CL). However, existing PCPR models often suffer from catastrophic forgetting, leading to significant performance degradation in previously learned scenes when adapting to new environments or sensor types. This results in poor model scalability, increased maintenance costs, and system deployment difficulties, undermining the practicality of PCPR. To address these issues, we propose LifelongPR, a novel continual learning framework for PCPR, which effectively extracts and fuses knowledge from sequential point cloud data. First, to alleviate the knowledge loss, we propose a replay sample selection method that dynamically allocates sample sizes according to each dataset's information quantity and selects spatially diverse samples for maximal representativeness. Second, to handle domain shifts, we design a prompt learning-based CL framework with a lightweight prompt module and a two-stage training strategy, enabling domain-specific feature adaptation while minimizing forgetting. Comprehensive experiments on large-scale public and self-collected datasets are conducted to validate the effectiveness of the proposed method. Compared with the state-of-the-art (SOTA) method, our method achieves 6.50\% improvement in $mIR@1$, 7.96\% improvement in $mR@1$, and an 8.95\% reduction in $F$. The code and pre-trained models are publicly available at \url{https://zouxianghong.github.io/LifelongPR}.
\end{abstract}


\section{Introduction}\label{sec_introduction}
\IEEEPARstart{P}{lace} recognition is a foundational task in geoscience and robotics, enabling autonomous systems to determining their geo-locations within previously mapped environments by identifying revisited places \citep{zhou2025whu,li2025saliencyi2ploc}. It serves as a key module in a wide range of applications, including autonomous driving \citep{yurtsever2020survey,li2023whu}, intelligent transportation systems \citep{ounoughi2023data,li2024hcto}, collaborative mapping \citep{xu2025atcm,buehrer2018collaborative}, and augmented reality \citep{carmigniani2011augmented,li2025ua}. In recent years, point cloud place recognition (PCPR) has gained increasing attention due to its robustness against illumination changes, occlusions, and seasonal variations. With the advancement of deep learning techniques, PCPR has achieved remarkable progress in both recognition accuracy and robustness \citep{uy2018pointnetvlad, komorowski2021minkloc3d, zou2023patchaugnet}, making it a promising solution for long-term and all-weather localization in complex environments.

\begin{figure*}
    \centering
    \includegraphics[width=1\linewidth]{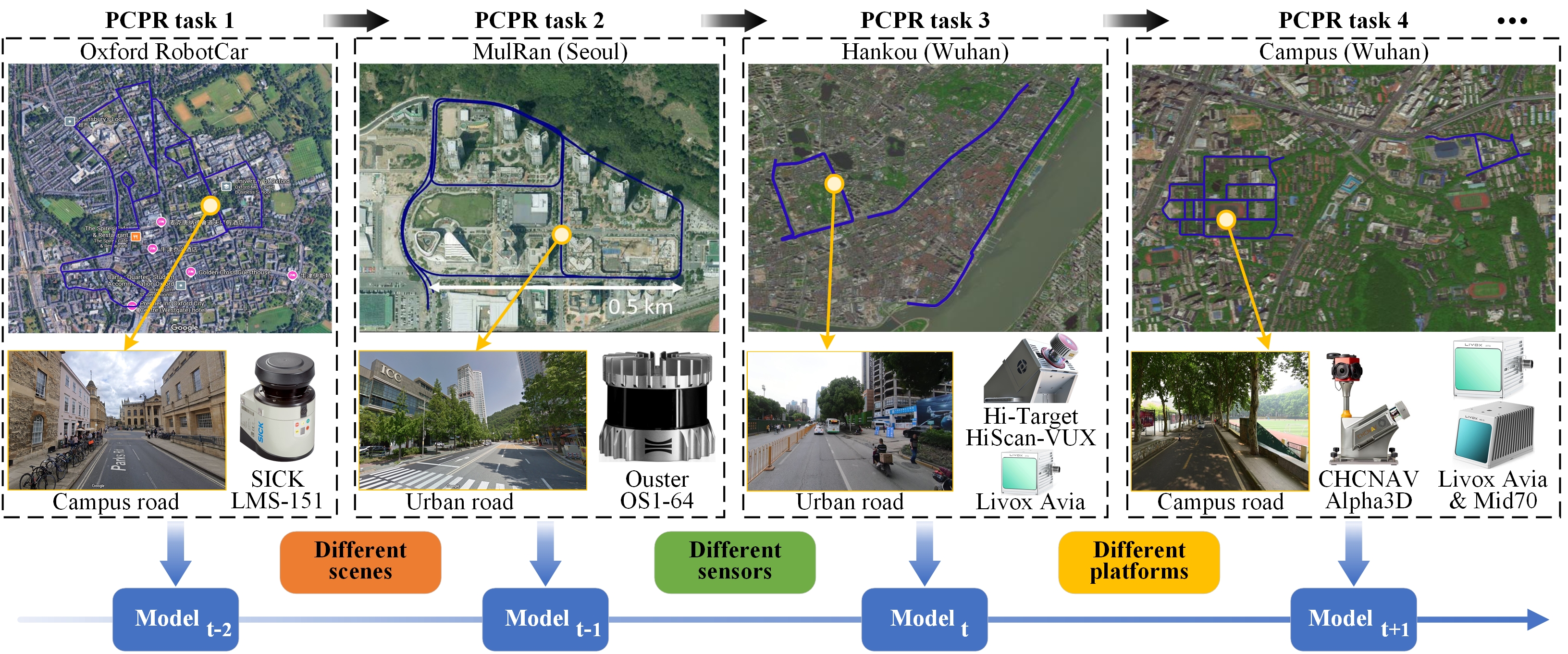}
    \caption{Lifelong point cloud place recognition encounters challenges posed by diverse localization scenes and LiDAR sensors. Above information mainly comes from the experimental datasets in Section \ref{sec_experiment}.}
    \label{fig:LifelongPR-why-continual-learning}
\end{figure*}

Despite this progress, real-world large-scale deployments of PCPR introduce new challenges. PCPR models are often required to operate across diverse scenes (urban, suburban, or campus) and under different LiDAR sensor types as shown in Fig. \ref{fig:LifelongPR-why-continual-learning}. Point clouds collected under such heterogeneous conditions exhibit significant domain shifts, which cause severe performance degradation when the model is applied beyond its original training data \citep{jung2024helipr}. To ensure consistent performance, the PCPR model must be continually updated with new data collected over time and across domains \citep{lesort2020continual}. However, such continual adaptation leads to catastrophic forgetting, where learning new environments overwrites knowledge of previously seen ones, undermining model reliability and safety. For instance, a model fine-tuned on a new city or sensor may lose up to 30\% recall on earlier deployment zones (see column 1 in Fig. \ref{fig:LifelongPR-comparison-res-dcc} and Fig. \ref{fig:LifelongPR-comparison-res-hete}). A naive solution would be to retrain the model jointly with all data, which is computationally expensive and unscalable. Alternatively, maintaining separate models for each scenario drastically increases storage and system complexity \citep{aljundi2017expert,serra2018overcoming}. These limitations highlight the urgent need for effective knowledge fusion frameworks that can incrementally learn from new data while retaining prior knowledge, enabling scalable and robust PCPR in real-world, lifelong applications.

Continual learning (CL) studies the problem of fusing knowledge from dynamic data distribution, aiming at mitigating catastrophic forgetting \citep{lesort2020continual,zheng2025enhancing}. In recent years, a few studies have introduced the theory of CL to mitigate catastrophic forgetting in PCPR \citep{knights2022incloud,cui2023ccl}.
These methods primarily rely on two strategies: sample replay and knowledge distillation. Replay-based methods store a limited number of past samples and interleave them with new data during training, while distillation-based approaches transfer knowledge from previous models by minimizing output divergence\citep{rebuffi2017icarl,li2017learning}.
While these approaches offer partial mitigation, they suffer from two fundamental limitations that significantly constrain their effectiveness in real-world deployments. First, existing methods typically adopt random sampling for replay memory, ignoring the information content and spatial distribution of training data. This leads to unrepresentative replay sets that fail to preserve essential scene knowledge, especially in large and diverse environments. Second, when domain gaps between training stages are substantial (e.g., due to changes in environment or LiDAR sensor), a small replay buffer becomes insufficient to retain old knowledge, while the effectiveness of knowledge distillation also declines due to increased feature drift.

To address the limitations of existing CL methods for PCPR, namely, unrepresentative replay samples and poor adaptability to domain shifts, we propose LifelongPR, a novel continual learning framework that enables effective knowledge fusion across sequential 3D domains. Specifically:

(1) We propose a replay sample selection strategy that addresses the problem of low replay efficiency. By dynamically allocating sample sizes based on the information quantity of each training set, and selecting samples with diverse spatial distribution, our method ensures that a small number of stored samples can better preserve prior knowledge. This significantly improves performance under strict memory constraints.

(2) We introduce a prompt learning-based CL framework to tackle the challenge of domain shift and feature drift in continual adaptation. A lightweight prompt module is used to capture domain-specific knowledge from each training set. The combination of a two-stage training strategy guides the backbone network to extract sample-adaptive features. This design enables the model to maintain place recognition performance across heterogeneous scenes without retraining or maintaining multiple models.

The remainder of this article is organized as follows. Section \ref{sec_related_work} reviews related works. Section \ref{sec_method} details the proposed method. Section \ref{sec_experiment} presents experimental data, and quantitative evaluation. Section \ref{sec_analysis_discussion} presents the ablation studies and discusses the deficiencies and future work. Finally, Section \ref{sec_conclusion} concludes the paper.

\section{Related Work}\label{sec_related_work}
\subsection{Learning-based point cloud place recognition}
Learning-based PCPR methods can be categorized into projection-based, voxel-based, and point-based according to the data type of input for neural networks.

\textbf{Projection-based methods} extract features from 2D images (e.g., range images and bird's eye view (BEV) images) generated from raw point clouds. OREOS \citep{schaupp2019oreos} feeds range images generated from point clouds into a CNN network and extracts features for characterizing position and orientation respectively, enabling simultaneous place recognition and yaw estimation. \citet{kim2019sci-loc} generates BEV images from point clouds based on Scan Context \citep{kim2018scan} and regards the PCPR as a classification task. OverlapNet \citep{chen2021overlapnet} feeds range images to a Siamese neural network and directly regresses the overlap ratio and yaw angle of two point clouds. While computationally efficient, these methods incur significant information loss in the process of converting point clouds into 2D images.

\textbf{Voxel-based methods} extract features from voxelized point clouds. Minkloc3D \citep{komorowski2021minkloc3d} proposes the first voxel-based PCPR method, extracting local and contextual features via feature pyramid and sparse convolution, while aggregating global features with GeM. \citet{komorowski2022improving} enhances it with the ECA attention \citep{wang2020eca} module and average precision loss. SVT-Net \citep{fan2022svt} combines the sparse convolution and transformer to extract global features, making the model lightweight and efficient. Such methods have simple network structures and are more noise resistant, while voxelization suffers from information loss.

\textbf{Point-based methods} extract features directly from raw point clouds. \citet{uy2018pointnetvlad} propose the first learning-based PCPR method built upon PointNet \citep{qi2017pointnet} and NetVLAD \citep{arandjelovic2016netvlad}. \citet{hui2021pyramid} utilizes the EdgeConv and UNet-like structure to enhance the feature representation ability of the network. \citet{zhou2021ndt} feeds probabilistic representation converted from point clouds into the network, increasing the noise immunity. PatchAugNet \citep{zou2023patchaugnet} enhances the adaptability of the network to domain gaps between heterogeneous point clouds with a patch feature augmentation module and an adaptive pyramid feature aggregation module. LAWS \citep{xie2024look} treats place recognition as a classification task and eliminates classification ambiguity through two spatial partitioning patterns. Such methods retain the information of original point clouds, but they have poor generalization ability to unseen scenes.

In general, existing learning-based PCPR methods suffer from severe overfitting on public datasets and have poor transferability to unseen scenes and sensor configurations.

\subsection{Continual learning}
CL, also termed incremental learning or lifelong learning, learns from dynamic data distributions, and its key problem is the trade-off between learning plasticity and memory stability \citep{wang2024comprehensive}. Existing CL methods primarily fall into three categories: replay-based, regularization-based, and parameter isolation-based \citep{de2021continual}.

\textbf{Replay-based methods} first keep original samples from previous tasks or generate pseudo samples from a generative model and then learn jointly with current task data. The most representative method is iCaRL \citep{rebuffi2017icarl}. Aiming at class incremental learning tasks, it selects replay samples according to the distance between replay samples and class centers and applies knowledge distillation to mitigate catastrophic forgetting. BIC \citep{wu2019large} utilizes class-balanced validation sets to train deviation parameters for network output adjustment, addressing the problem of imbalanced new and replay sample sizes in class incremental learning tasks. DGR \citep{shin2017continual}, the first CL method utilizing the generative model, alternately trains the task network and generative network. MeRGAN \citep{wu2018memory} further utilizes replay alignment to enhance the consistency between old and new generative models, playing a role similar to regularization. Such methods usually mitigate catastrophic forgetting well, and its key point lies in replay sample selection and utilization \citep{macqueen1967some,killamsetty2021grad,kothawade2022prism}.

\textbf{Regularization-based methods} balance current and previous tasks through explicit regularization items, falling into two categories: weight regularization-based and function regularization-based. Weight regularization-based methods selectively regularize the variation of network parameters by evaluating the importance of network parameters to previous tasks \citep{kirkpatrick2017overcoming}. EWC \citep{kirkpatrick2017overcoming} is the most representative weight regularization-based method. Function regularization-based methods utilize the previous models to guide the training of the current model through knowledge distillation. LwF \citep{li2017learning} first introduces knowledge distillation to CL and regards the output of previous models as soft labels for training the current model. Such methods are simple and effective but sensitive to domain gaps between tasks.

\textbf{Parameter isolation-based methods} allocate dedicated parameters for specific tasks and can guarantee maximal stability by fixing the parameter subsets of previous tasks. Methods like HAT \citep{serra2018overcoming} and PackNet \citep{mallya2018packnet} retain multiple task-specific models with identical network structures. Other methods like Expert Gate \citep{aljundi2017expert} and PathNet \citep{fernando2017pathnet} divide the network into task-sharing and task-specific parts and limit the interference between tasks to the task-specific network structure. Such methods avoid or mitigate catastrophic forgetting via task-specific networks, while they must maintain multiple models.

In recent years, the above CL methods have been introduced to PCPR for catastrophic forgetting mitigation. InCloud \citep{knights2022incloud}, the first CL method for PCPR, utilizes sample replay and knowledge distillation to mitigate catastrophic forgetting. CCL \citep{cui2023ccl} combines contrastive learning and CL to improve the network's generalizability and mitigate catastrophic forgetting. Both methods' random replay sampling fails to adequately represent historical training data. BioSLAM \citep{yin2023bioslam} achieves CL for cross-modal place recognition via GAN-based sample generation and dual-memory replay selection, but incurs increased training complexity and reduced efficiency.

\subsection{Prompt learning for continual learning}
Prompt learning is an emerging technique, designing task-specific prompts to conditionally adapt pre-trained models for downstream tasks \citep{liu2023pre}. Unlike Adapter \citep{houlsby2019parameter} and LoRA \citep{hu2021lora}, it avoids architectural modifications with minimal parameter additions. Some methods strike a balance between pre-trained knowledge and downstream tasks with lightweight prompts, serving as external memory for CL \citep{zhou2024continual}. L2P \citep{wang2022learning} is the first method to introduce prompt learning to CL. It proposes a learnable prompt pool to instruct the pre-trained model to perform downstream tasks conditionally. DualPrompt \citep{wang2022dualprompt} proposes dual prompts for CL. G-prompt and E-prompt learn task-invariant and task-specific instructions respectively. S-Prompt \citep{wang2022s-prompt}  builds task centers by K-means clustering and utilizes the KNN search to find the most similar prompt. DAP \citep{jung2023generating} utilizes a lightweight MLP network to construct instance-level prompts to resolve domain scalability issues. Existing methods are predominantly designed for transformer-based networks, and how to use prompts for non-transformer networks still need further research.

\section{Methodology}\label{sec_method}
\begin{figure*}[htbp]
    \centering
    \includegraphics[width=1\linewidth]{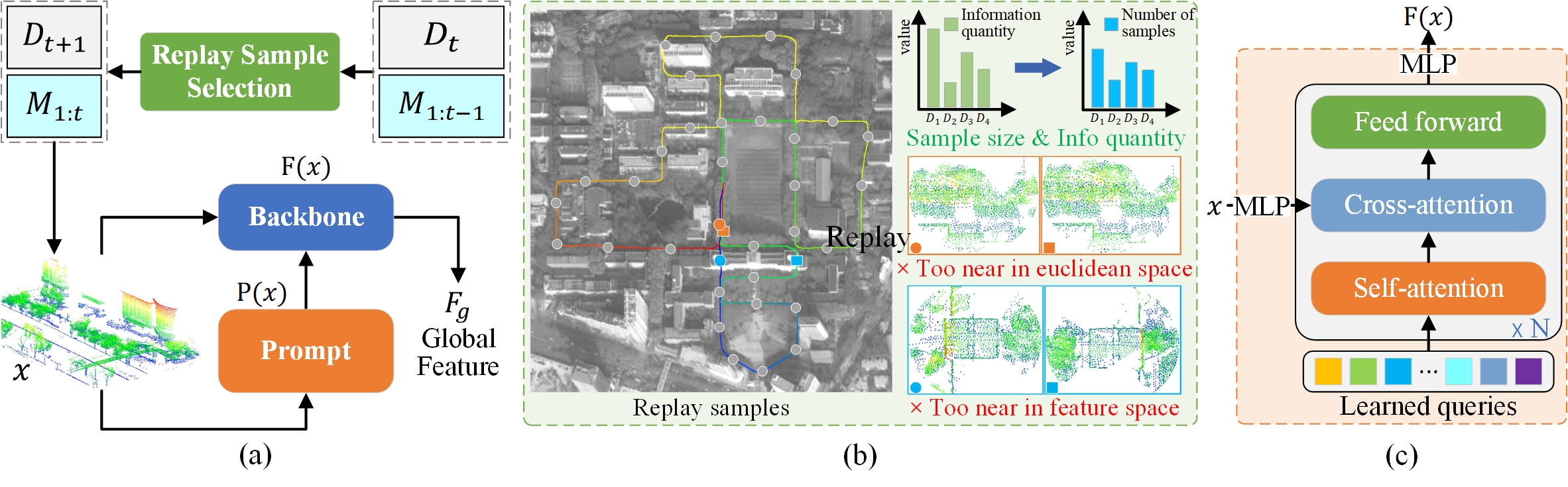}
    \caption{Overview of the proposed method. (a) overall workflow, (b) replay sample selection, (c) prompt module.}
    \label{fig:LifelongPR-method}
\end{figure*}

Our method consists of 1) a replay sample selection method considering information quantity and spatial distribution, and 2) a CL framework based on prompt learning, as shown in Fig. \ref{fig:LifelongPR-method}. Firstly, the proposed replay sample selection method dynamically allocates a replay sample size to each  training set based on information quantity and selects replay samples based on spatial distribution. Secondly, combined with a two-stage training strategy, the lightweight prompt module is trained to extract domain-specific knowledge of each training set and guide the backbone network to extract sample-adaptive features.

\subsection{Definition of replay-based CL for PCPR}
\begin{figure}
    \centering
    \includegraphics[width=1\linewidth]{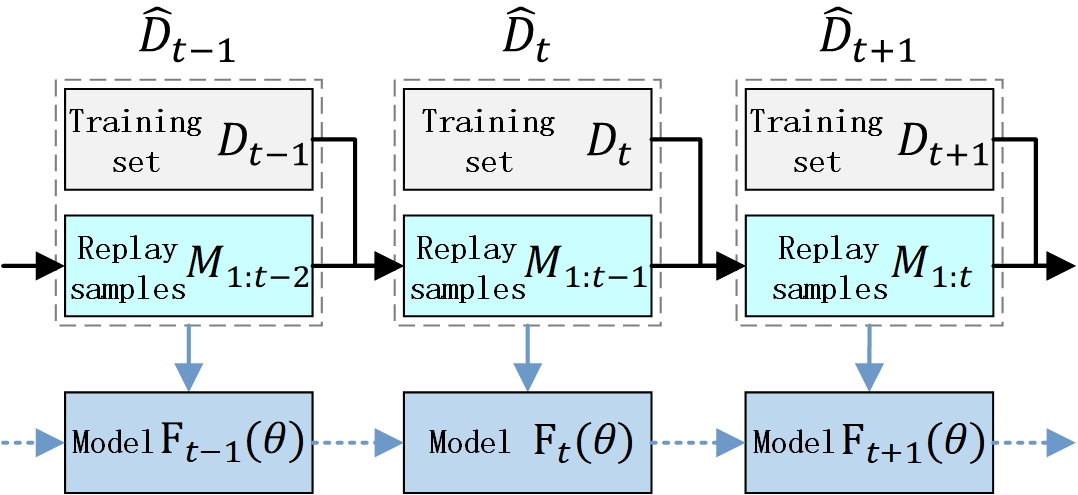}
    \caption{Replay-based continual learning for point cloud place recognition.}
    \label{fig:LifelongPR-Incremental-PR}
\end{figure}

As shown in Fig. \ref{fig:LifelongPR-Incremental-PR}, replay-based CL for PCPR is a domain-incremental learning task, which uses $T$ discontinuous training sets $D_{1:T}=\{ D_t \}^T_{t=1}$ to update the PCPR model $\mathrm{F}(\theta)$ \footnote{The following section omits the symbol $\theta$. } incrementally, aiming to achieve better place recognition performance in their respective scenes. Each training set $D_t$ is collected from different scenes or using different LiDAR sensors, leading to domain gaps among various training sets. When training on $D_t$, the complete historical training sets $D_{1:t-1}$ cannot be fully accessed; instead, only a subset $M_{1:t-1} \subset D_{1:t-1}$ can be retrieved for replay. Given constraints on training efficiency and limited storage resources, the total number of replay samples $k_{total}$ is typically fixed. After updating the place recognition model with the current training set and the replay samples $\hat{D}_t=D_t \cup M_{1:t-1}$, the model's performance still tends to decline in the scenes corresponding to the historical training sets $D_{1:t-1}$, i.e. catastrophic forgetting. This is the core challenge in CL for PCPR.

\subsection{Replay sample selection considering information quantity and spatial distribution}
Sample replay is a simple and effective method to mitigate catastrophic forgetting in CL. The key challenge lies in selecting samples that best represent historical training sets. It involves solving two key problems: 1) \textit{How many samples to select from each training set}; 2) \textit{Which samples to select for replay}. To solve these two problems, we propose a replay sample selection method considering the information quantity of training sets and the spatial distribution of samples, as shown in Fig. \ref{fig:LifelongPR-method} (b).

\subsubsection{Sample size allocation based on information quantity}
Existing CL methods for PCPR almost uniformly select the same number of samples from all training sets for replay, overlooking variations in information quantity across training sets. To this end, we quantitatively assess the information quantity of training sets and, dynamically allocating the number of replay samples to select from each training set.

\begin{remark}
The information quantity of a training set is related to factors such as the size and diversity of samples, providing guidance on how many samples to select from each training set. The rank of the kernel matrix \citep{fine2001efficient} represents the maximum number of linearly independent samples in the training set and thus reflects the magnitude of its information content, enabling quantitative assessment of the training set's information quantity.
\end{remark}

\textbf{Information quantity}. Given a training set $D$, where the features of its $n$ samples constitute the set $\{f_i\}$, we construct the corresponding kernel matrix $A \in R^{n \times n}$ using the Gaussian kernel function as follows:
\begin{equation}\label{eq:kernel-matrix}
    A_{ij}=\exp(-\gamma||f_i-f_j||^2), i,j=1,\cdots,n,
\end{equation}
where $\gamma$ is a parameter of the Gaussian kernel function, and $f_i$ is the feature of the $i$th sample in $D$. Then, we use singular value decomposition (SVD) to estimate the effective rank of $A$ as follows:
\begin{equation}\label{eq:SVD}
    A=U \Sigma V^T, \Sigma = \mathrm{diag}(\sigma_1, \sigma_2,\cdots,\sigma_n),
\end{equation}
\begin{equation}\label{eq:rank}
    \mathrm{Rank}(A)=\# \{\sigma_i|\sigma_i \geq \epsilon \cdot \sigma_1\},
\end{equation}
where $\epsilon$ is a empirical parameter. Then, we normalize the effective rank to $[0,1]$ to quantitatively measure the account of information in training set $D$ as follows:
\begin{equation}\label{eq:normal-rank}
    \mathrm{InfoQ}(D)=\frac{\mathrm{Rank}(A)}{n}.
\end{equation}
The closer the value of $\mathrm{InfoQ}(D)$ is to 1, the greater the sample diversity and information quantity of training set $D$. For convenience, we denote $\mathrm{InfoQ}(D_t)$ as $InfoQ_t$.

\textbf{Sample size allocation}. Given the training set sequence $D_{1:t}$ and the fixed total number of replay samples $k_{total}$, the corresponding information quantity $InfoQ_{1:t}$ is calculated via Eq. (\ref{eq:kernel-matrix}-\ref{eq:normal-rank}). The number of replay samples $k_t$ allocated to training set $D_t$ is calculated using a temperature-controlled softmax function as follows:
\begin{equation}\label{eq:sample_allocation}
    k_t=k_{total} \cdot \frac{\exp(\mathrm{InfoQ}(D_t)/\tau)}{\Sigma^{T}_{i=1}\exp(\mathrm{InfoQ}(D_i)/\tau)}, \tau \in (0,\infty),
\end{equation}
where $\tau$ is the temperature threshold, and when $\tau \rightarrow \infty$, Eq. (\ref{eq:sample_allocation}) allocates the same number of replay samples to each training set. According to Eq. (\ref{eq:sample_allocation}), we can obtain the replay sample sizes $k_{1:t}$ corresponding to $D_{1:t}$.

\subsubsection{Sample selection based on spatial distribution}
Given training set $D$, $k$ samples are selected to form a set $M$ such that $M$ best represents $D$, i.e., the sample diversity of $M$ is maximized. Specifically, the problem can be expressed as:
\begin{equation}\label{eq:maximize_gM}
    M^* = \underset{M \subset D}{\operatorname{arg\,max} \, \mathrm{g}(M)}, s.t., |M| = k,
\end{equation}
where $M^*$ is the optimal subset, $\mathrm{g}(M)$ represents the sample diversity of $M$, and it is constructed upon spatial distribution.

\textbf{Spatial distribution}. When selecting replay samples, we take into account their distributions in both Euclidean and feature spaces. For each sample $x_i \in M$, the minimum distances between $x_i$ and other samples in $M$ in Euclidean and feature spaces are calculated as follows:
\begin{equation}\label{eq:spatial_distribution}
    S^i=\min _{j \in M, j \neq i} (\min \left(d_{i, j}/d_{thr}, 1\right) + (1-\cos \left(f_{i}, f_{j}\right))/2),
\end{equation}
where $d_{i, j}$ is the distance between two samples $x_i$ and $x_j$, $d_{thr}$ is a threshold of Euclidean distance, and $f_i$ is the feature of $x_i$. Based on spatial distribution of samples, we construct a function $\mathrm{g}(M)$ to quantitatively measure the diversity of samples in $M$. A larger $\mathrm{g}(M)$ indicates greater sample diversity of $M$, which means it contains more information and is more representative of the original training set $D$. The definition of $\mathrm{g}(M)$ is as follows:
\begin{equation}\label{eq:gM_def}
    \mathrm{g}(M) = \Sigma_{i \in M} S^i.
\end{equation}

\textbf{Sample selection}. Eq. (\ref{eq:maximize_gM}) prevents the selected samples from being overly concentrated in both Euclidean and feature spaces, but it's an NP-hard problem. Hence we use efficient greedy algorithm to solve Eq. (\ref{eq:maximize_gM}), i.e. select the most representative $k$ samples from the training set $D$. In each greedy searching step, we randomly select samples (the set is denoted as $R$) from $D \backslash M$ to speed up the searching process. The definition of $R$ is:
\begin{equation}\label{eq:num_R}
    R \subseteq D \backslash M, |R|=\frac{|D|}{k} \alpha,
\end{equation}
where $\alpha$ is a ratio threshold. The time complexity of the greedy algorithm is $O(\alpha |D|)$, and is independent of the selected sample size.

Given the number of replay samples $k^{\prime}_{1:t-1}$ and replay sample set sequence $M^{\prime}_{1:t-1}$ corresponding to the training set sequence $D_{1:t-1}$. When learning on the $t$th training set $D_t$, the replay sample sizes $k_{1:t}$ of the training set sequence $D_{1:t}$ can be obtained via Eq. (\ref{eq:sample_allocation}), and the corresponding replay sample set sequence is updated to $M_{1:t}$. First, we select a new replay sample set $M_t$ (where $|M_t|=k_t$) from $D_t$, i.e. , sample selection. Second, we remove some samples from the original replay sample set sequence $M^{\prime}_{1:t-1}$ to obtain the new replay sample set sequence $M_{1:t-1}$, i.e., sample forgetting. For more details, see Algorithm \ref{algo:sample_selection}.

\begin{algorithm}
\small
\SetAlgoLined 
\SetKwProg{Fn}{Function}{:}{end}
\SetKwComment{Comment}{// }{}
\caption{Replay Sample Selection}
\label{algo:sample_selection}

\KwIn{$M^{\prime}_{1:t-1}$, $InfoQ_{1:t-1}$, $D_t$, $k_{total}$, $\alpha$.}
\KwOut{$M_{1:t}$.}

\Fn{GreedySelect($D$, $k$, $\alpha$)}
{
$M \leftarrow \emptyset$\;
\For{$i \leftarrow 1$, $i \leq k$, $i \leftarrow i+1$}
{
Randomly select $R \leftarrow D \backslash M$ via Eq. (\ref{eq:num_R})\;
$m_i \leftarrow \max_{m_i \in R} \mathrm{g}(M \cup \{m_i\})$ via Eq. (\ref{eq:spatial_distribution}-\ref{eq:gM_def})\;
$M \leftarrow M \cup \{m_i\}$\;
}
\Return{$M$}
}

\Fn{SelectSamples($M_{1:t-1}$, $InfoQ_{1:t-1}$, $D_t$, $k_{total}$)}
{
\Comment{allocate sample size for each training set}
Calculate $InfoQ_t$ of $D_t$ via Eq. (\ref{eq:kernel-matrix}-\ref{eq:normal-rank})\;
Allocate sample sizes $k_{1:t}$ via Eq. (\ref{eq:sample_allocation})\;
\Comment{select new replay samples}
$M_t=\mathrm{GreedySelect}(D_t, k_t, \alpha)$\;
\Comment{forget some history replay samples}
\For{$i \leftarrow 1$, $i \leq t-1$, $i \leftarrow i+1$}
{
$M_i=\mathrm{GreedySelect}(M^{\prime}_i, k_i, \alpha)$\;
}
$M_{1:t}=M_{1:t-1} \cup \{M_t\}$\;
\Return{$M_{1:t}$}
}
\end{algorithm}

\subsection{CL framework based on prompt learning}
Prompt learning is a novel transfer learning strategy that adapts pre-trained models to new tasks or scenarios by utilizing learnable prompts. It effectively preserves the existing knowledge of the pre-trained model while adapting to new tasks, offering high training efficiency and excellent transfer performance. CL involves learning from dynamic data distributions, requiring adaptation to new data distributions while retaining knowledge acquired from old ones. Therefore, this section introduces prompt learning into CL for PCPR, with the workflow illustrated in Fig. \ref{fig:LifelongPR-method} (c).

\textbf{Prompt module.} We design the prompt module with reference to QFormer \citep{li2023blip}. It consists of two MLPs (denoted as $\mathrm{MLP_{in}}:R^3 \rightarrow R^d$ and $\mathrm{MLP_{out}}: R^d \rightarrow R^{d'}$), $k_q$ learnable prompts (denoted as $Q \in R^{k_q \times d}$), and $N$ attention modules (denoted as $\mathrm{Attn}: R^d \rightarrow R^d$), as shown in Fig. \ref{fig:LifelongPR-method} (c). Here, $\mathrm{MLP_{in}}$ maps the raw point cloud (denoted as $x \in R^{N_x \times 3}$) into a high-dimensional space, which then serves as the input to the attention modules. The learnable prompts $Q$ store domain-specific information from each training set and act as the Query in the attention modules. The attention modules $\mathrm{Attn}$ facilitate interaction between $\mathrm{MLP_{in}}(x)$ and $Q$ to extract domain-specific information related to $x$. $\mathrm{MLP_{out}}$ adjusts the dimensionality of the attention module outputs, making it easier to embed prompt information into the backbone network. The dimension $d'$ is tailored to accommodate different backbone networks. The extraction of domain-specific information from raw point cloud $x$ by the prompt module is formally expressed as:
\begin{equation}
    \mathrm{P}(x)=\mathrm{MLP_{out}}(\mathrm{Attn}(\mathrm{MLP_{in}}(x),Q)) \in R^{N_x \times d'.}
\end{equation}
After incorporating the prompt module, the updated PCPR model $\mathrm{H}_t = \{\mathrm{F}_t, \mathrm{P}_t\}$ is obtained. Global feature extraction from raw point cloud $x$ is then formalized as: 
\begin{equation}
    \mathrm{H}_t(x)=\mathrm{F}_t(x,\mathrm{P}_t(x)).
\end{equation}

\begin{figure*}[b]
    \centering
    \includegraphics[width=0.95\linewidth]{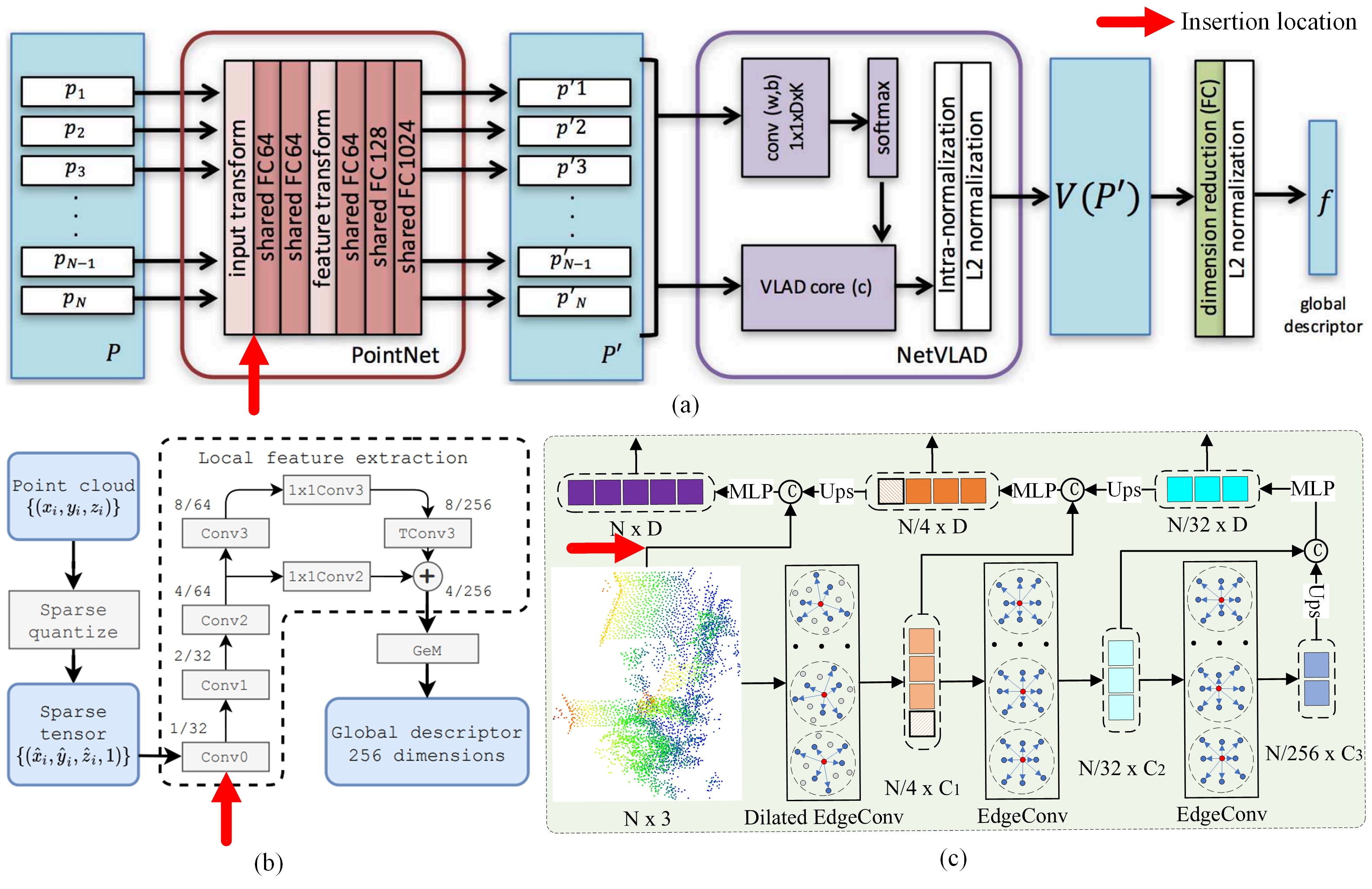}
    \caption{Location of the prompt module within the backbone network. (a) PointNetVLAD, (b) Minkloc3D, (c) PatchAugNet. Red arrows denote the insertion locations of of the prompt module.}
    \label{fig:LifelongPR-prompt-location}
\end{figure*}

Determining how to integrate learnable prompts into the backbone network is a critical problem. Existing prompt learning methods are primarily designed for transformer-based networks. However, most deep learning-based PCPR methods rely on sparse convolution or PointNet-style architectures, requiring the insertion location of prompts to be tailored to the specific structure of the backbone network. For Minkloc3D, a PCPR method based on sparse convolution, this study directly serializes $\mathrm{P}(x)$ with the raw point cloud and uses it as the network input. For PCPR models based on PointNet, such as PointNetVLAD and PatchAugNet, $\mathrm{P}(x)$ is directly added to the intermediate features of the backbone network. Crucially, our method requires no modifications to backbone networks. The insertion locations of the prompt module are visualized in Fig. \ref{fig:LifelongPR-prompt-location}.

\textbf{Training strategy.} Existing prompt learning methods typically adapt pre-trained models to downstream tasks via learnable prompts without updating parameters of the pre-trained model. However, small models, such as PointNetVLAD and Minkloc3D, have limited feature representation and generalization capabilities, requiring parameter updates when adapting to new domains with prompt learning. To ensure minimal adjustments to the backbone network parameters during CL while enabling the prompt module to capture domain-specific information from each training set, we propose a two-stage training strategy. In stage 1, backbone parameters remain frozen while the prompt module trains with replay samples to extract domain-specific knowledge. In stage 2, the prompt module freezes while the backbone fine-tunes on the current training set. Both stages maintain identical optimization objectives throughout this process.
\begin{figure}[h]
    \centering
    \includegraphics[width=1\linewidth]{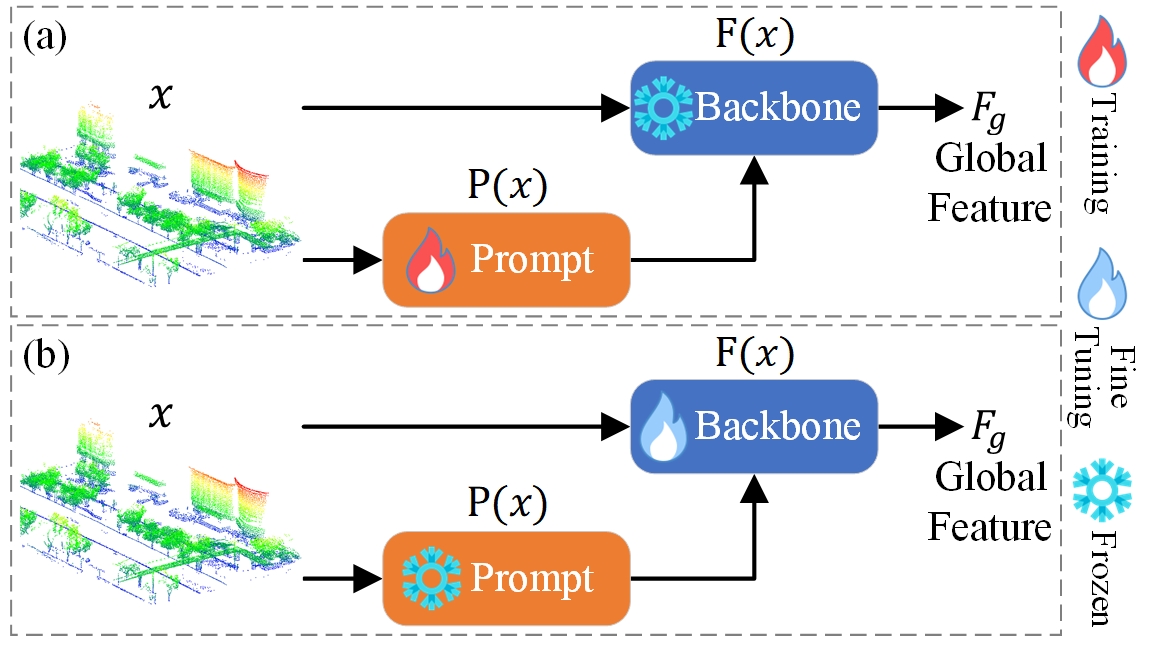}
    \caption{Two-stage training strategy. (a) stage 1, (b) stage 2.}
    \label{fig:LifelongPR-training-strategy}
\end{figure}

\subsection{Loss function}
The loss function consists of two parts as follows:
\begin{equation}
    L = L_{PR} + \lambda ^ \gamma L_{KD},
\end{equation}
where $L_{PR}$ is the triplet loss, used to supervise the place recognition task. $L_{KD}$ is the knowledge distillation loss, designed to ensure that the outputs of replay samples remain consistent between the updated model and the historical model. $\lambda ^ \gamma$ is the weight of $L_{KD}$, $\gamma$ is epoch index, and $\lambda ^ \gamma$ monotonically decreases with $\gamma$. For more details, please refer to InCloud \citep{knights2022incloud}.

\section{Experiment}\label{sec_experiment}
\subsection{Datasets and settings}\label{subsec_exp_data_setting}

\textbf{Datasets.}
The experimental data encompasses large-scale public datasets (Oxford RobotCar \citep{maddern20171}, MulRan \citep{kim2020mulran}, and In-house \citep{uy2018pointnetvlad}) and our self-collected large-scale heterogeneous point cloud dataset (refer to PatchAugNet \citep{zou2023patchaugnet}). Specifically, MulRan consists of two parts: DCC and Riverside, while the heterogeneous point cloud dataset comprises two sections: Hankou and Campus. These datasets were collected in different cities using various mounting platforms and LiDAR sensors. All experimental data have undergone ground removal and coordinate normalization, with each point cloud submap containing 4,096 points. We incrementally train the PCPR models on two dataset sequences: Seq1 (Oxford$\rightarrow$DCC$\rightarrow$Riverside$\rightarrow$In-house) and Seq2 (Oxford$\rightarrow$Hankou$\rightarrow$Campus$\rightarrow$In-house) respectively. Compared with Seq1, Seq2 exhibits greater data distribution differences, i.e. larger domain gap. More details are shown in Table \ref{tab:exp_data}.

\begin{table*}
\centering
\caption{Details of the experimental data.}
\label{tab:exp_data}
\begin{tabular}{ccccc} 
\hline
\multirow{2}{*}{\textbf{Dataset}} & \multirow{2}{*}{\textbf{ Scene}}                   & \multirow{2}{*}{Platform}                               & \multirow{2}{*}{\textbf{LiDAR sensor}}                                          & \textbf{Num of submaps}         \\
                                   &                                                    &                                                         &                                                                                  & \textbf{train /~\textbf{test}}  \\ 
\hline
Oxford                             & Urban road (Oxford)~                               & Vehicle                                                 & SICK LMS-151                                                                     & 21711 / 3030                    \\
In-house                           & Urban road                                         & Vehicle                                                 & Velodyne HDL-64E                                                                 & 6671 / 1766                     \\
DCC                                & Urban road (\textcolor[rgb]{0.2,0.2,0.2}{Seoul})   & Vehicle                                                 & Ouster OS1-64                                                                    & 5542 / 15039                    \\
Riverside                          & Urban road (\textcolor[rgb]{0.2,0.2,0.2}{Seoul}) & Vehicle                                                 & Ouster OS1-64                                                                    & 5537 / 18633                    \\
Hankou                             & Urban road (Wuhan)                                 & \begin{tabular}[c]{@{}c@{}}Vehicle\\Helmet\end{tabular} & \begin{tabular}[c]{@{}c@{}}Livox Avia\\Hi-Target HiScan-VUX\end{tabular}         & 11329 / 2530                    \\
Campus                         & Campus road (Wuhan)                                & \begin{tabular}[c]{@{}c@{}}Vehicle\\Helmet\end{tabular} & \begin{tabular}[c]{@{}c@{}}Livox Mid70\\Livox Avia\\CHCNAV Alpha 3D\end{tabular} & 4020 / 1059                     \\
\hline
\end{tabular}
\end{table*}

\textbf{Baselines.}
We employ three representative methods for PCPR as backbone networks, namely PointNetVLAD \citep{uy2018pointnetvlad}, PatchAugNet\footnote{For convenience, the patch feature augmentation module is not used.} \citep{zou2023patchaugnet}, and Minkloc3D \citep{komorowski2021minkloc3d}. For CL baselines, we compare against Fine-tuning (denoted as FT), InCloud \citep{knights2022incloud}, and CCL \citep{cui2023ccl}. Among these, InCloud is the first CL method tailored for PCPR, mitigating catastrophic forgetting through sample replay and knowledge distillation. CCL represents the current SOTA (state-of-the-art) method, it mitigates catastrophic forgetting by introducing contrastive learning to enhance the network's generalization ability.

\textbf{Evaluation metrics.}
Referring to InCloud \citep{knights2022incloud}, we use the mean top-1 recall ($mR@1_k$) and forgetting score ($F$) to quantitatively evaluate the performance of each method in mitigating catastrophic forgetting. Here, $mR@1_k$ represents the average top-1 recall of the model on the seen datasets after completing the learning of the $k$th stage. $F$ is a quantitative assessment of the degree of forgetting of history knowledge after finishing training on all datasets. A lower value is preferable. Additionally, we introduce a new metric, i.e., mean incremental top-1 recall ($mIR@1_k$), which denotes the average top-1 recall in seen scenes during continual learning \citep{wang2024comprehensive}. Compared to $mR@1_k$, this metric comprehensively considers the performance of each learning stage. In the CL task for PCPR, after completing the $t$th learning stage, a new model $\mathrm{F}_t(x)$ and $\mathrm{P}_t(x)$ are obtained. The corresponding top-1 recall $Recall_{t,1:t}$ are acquired on the first $t$ test sets. Upon completing training on all $T$ datasets, a recall matrix $Recall$ is obtained. The specific definitions of three metrics are as follows:
\begin{equation}
    mR@1_k=\frac1k\sum_{t=1}^k Recall_{k,t},
\end{equation}
\begin{equation}
    mIR@1_k=\frac1k\sum_{t=1}^k mR@1_k,
\end{equation}
\begin{equation}
    F=\frac{1}{T-1}\sum_{t=1}^{T-1}\max_{l\in\{1,\cdots,T-1\}}\left\{Recall_{l,t}\right\}-Recall_{T,t}.
\end{equation}
In the following text, $mR@1$ and $mIR@1$ are used to refer to $mR@1_T$ and $mIR@1_T$ respectively.

\textbf{Implementation details.}
We set parameters for baseline methods by referencing the original papers and official code. The batch size is set based on GPU memory capacity. The thresholds $\gamma$, $\epsilon$, $\tau$, and $d_{thr}$ are set to 0.2, $10^{-6}$, 4.0, and $10^3$m respectively, and the total number of replay samples $k_{total}$ is 256. In our prompt module, the number of learnable prompts $k_q$ is 64, with a dimension $d$ of 8. The number of attention modules $N$ is 2, following the configuration of QFormer. The parameters for the loss function is consistent with InCloud. Our method employs the Adam optimizer with an initial learning rate of $10^{-3}$ and is trained for 40 epochs on each training set. When utilizing the prompt module, the model is trained for 40 epochs in both stage 1 and stage 2. The initial learning rate is set to $10^{-3}$ for stage 1 and $2 \times 10^{-4}$ for stage 2. All methods are implemented using PyTorch, and the experiments are conducted on a computer equipped with an NVIDIA GeForce RTX 4080S GPU.

\subsection{Quantitative evaluation}\label{subsec_quanitative_eval}
\begin{figure*}
    \centering
    \includegraphics[width=1\linewidth]{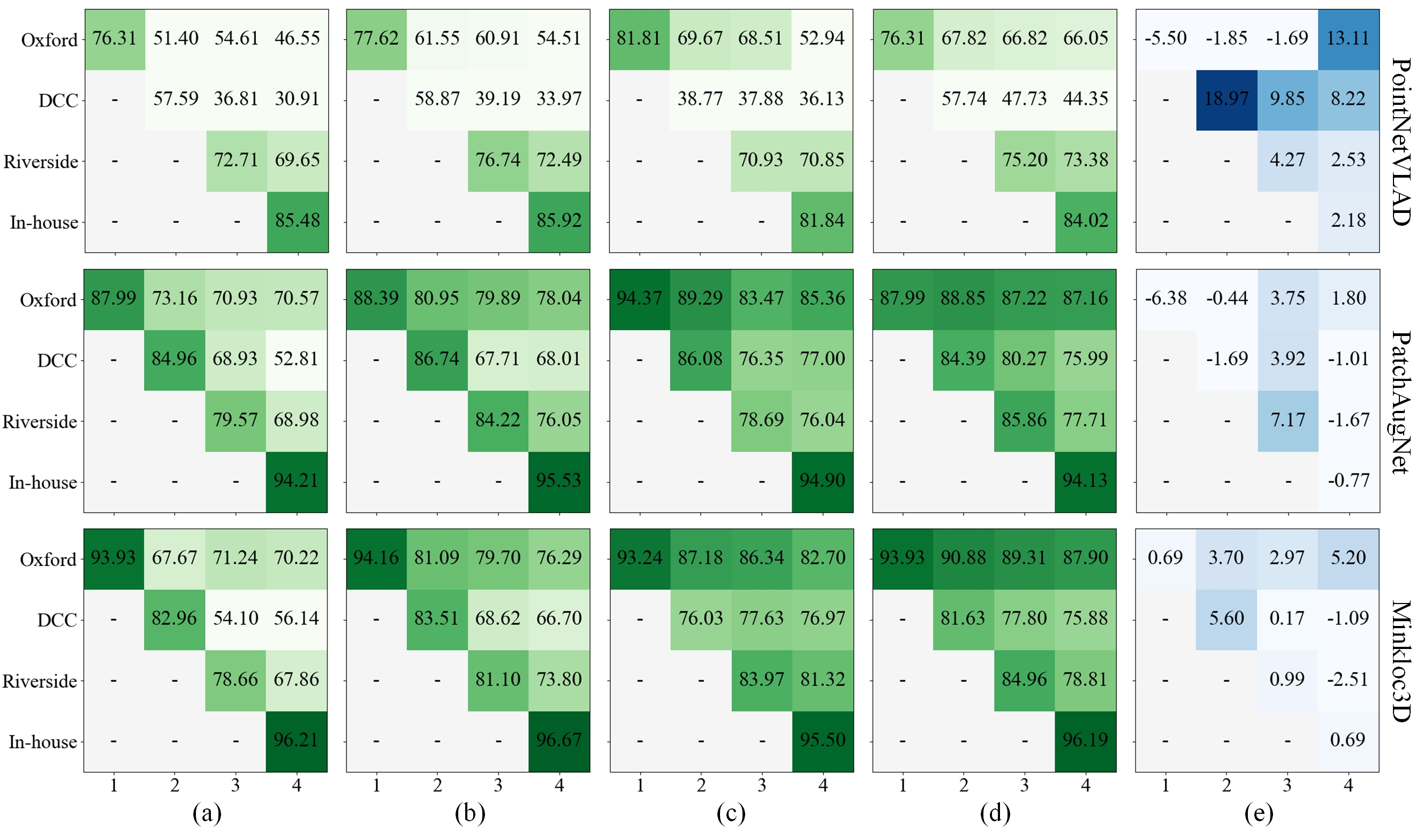}
    \caption{Average Recall@1 on Seq1. (a) FT, (b) InCloud, (c) CCL (SOTA method), (d) LifelongPR (ours), (d) improvements of LifelongPR compared to CCL. Rows 1-3 correspond to different backbone networks.}
    \label{fig:LifelongPR-comparison-res-dcc}
\end{figure*}

\textbf{Quantitative evaluation on two sequences.}
Table \ref{tab:comparison-res-seq1} and Fig. \ref{fig:LifelongPR-comparison-res-dcc} are the quantitative evaluation results of all methods on Seq1. As presented in Table \ref{tab:comparison-res-seq1} and Fig. \ref{fig:LifelongPR-comparison-res-dcc}, LifelongPR consistently achieves the best performance when incrementally trained on dataset sequence Seq1 with different backbone networks. Compared with the SOTA method CCL, LifelongPR demonstrates significant improvements, with a 3.43\% improvement in $mIR@1$, a 6.51\% improvement in $mR@1$, and a 2.04\% reduction in $F$. Notably, LifelongPR demonstrates particularly substantial performance improvement when employing PointNetVLAD as the backbone network. When employing Minkloc3D as the backbone network, LifelongPR achieves best CL results in terms of recall, surpassing CCL by 1.82\% in $mIR@1$, 0.57\% in $mR@1$. Although LifelongPR is 1.58\% lower than CCL on the $F$ metric, this occurs because CCL overfits on Oxford, degrading its initial performance on DCC and consequently leading to a smaller F.

\begin{table}
\centering
\fontsize{7.5}{10}\selectfont
\caption{Continual learning results on Seq1. The red and green numbers indicate the performance gains of our method over the SOTA approach (CCL).}
\label{tab:comparison-res-seq1}
\begin{tabular}{ccccc} 
\hline
\textbf{Backbone}             & \textbf{CL Method}  & \textbf{mIR@1(\%)$\uparrow$}                                                                   & \textbf{mR@1(\%)$\uparrow$}                                      & \textbf{F(\%)$\downarrow$}                                         \\ 
\hline
\multirow{4}{*}{PointNetVLAD} & FT                  & 60.92                                                                            & 58.14                                              & 19.84                                              \\
                              & InCloud             & \underline{64.62}                                                                            & \underline{61.72}                                              & 17.42                                              \\
                              & CCL                 & 63.89                                                                            & 60.44                                              & \underline{10.53}                                              \\
                              & \textbf{LifelongPR} & \textbf{67.32}\scalebox{0.7}{\textcolor[rgb]{0,0.86,0}{+3.43}} & \textbf{66.95}\scalebox{0.7}{\textcolor[rgb]{0,0.86,0}{+6.51}} & \textbf{8.49}\scalebox{0.7}{\textcolor[rgb]{0,0.86,0}{-2.04}}  \\ 
\hline
\multirow{4}{*}{PatchAugNet}  & FT                  & 77.96                                                                            & 71.64                                              & 20.06                                              \\
                              & InCloud             & 82.23                                                                            & 79.41                                              & 12.42                                              \\
                              & CCL                 & \textbf{86.22}                                                                   & \underline{83.32}                                              & \underline{6.91}                                               \\
                              & \textbf{LifelongPR} & \underline{85.70} \scalebox{0.7}{\textcolor{red}{-0.52}}                                                   & \textbf{83.75}\scalebox{0.7}{\textcolor[rgb]{0,0.86,0}{+0.42}} & \textbf{5.79}\scalebox{0.7}{\textcolor[rgb]{0,0.86,0}{-1.12}}  \\ 
\hline
\multirow{4}{*}{Minkloc3D}    & FT                  & 77.47                                                                            & 72.61                                              & 20.44                                              \\
                              & InCloud             & 82.83                                                                            & 78.36                                              & 14.00                                              \\
                              & CCL                 & \underline{85.40}                                                                            & \underline{84.12}                                              & \textbf{4.40}                                      \\
                              & \textbf{LifelongPR} & \textbf{87.23}\scalebox{0.7}{\textcolor[rgb]{0,0.86,0}{+1.82}}                               & \textbf{84.70}\scalebox{0.7}{\textcolor[rgb]{0,0.86,0}{+0.57}} & \underline{5.98} \scalebox{0.7}{\textcolor{red}{+1.58}}                      \\
\hline
\end{tabular}
\end{table}

\begin{figure*}
    \centering
    \includegraphics[width=1\linewidth]{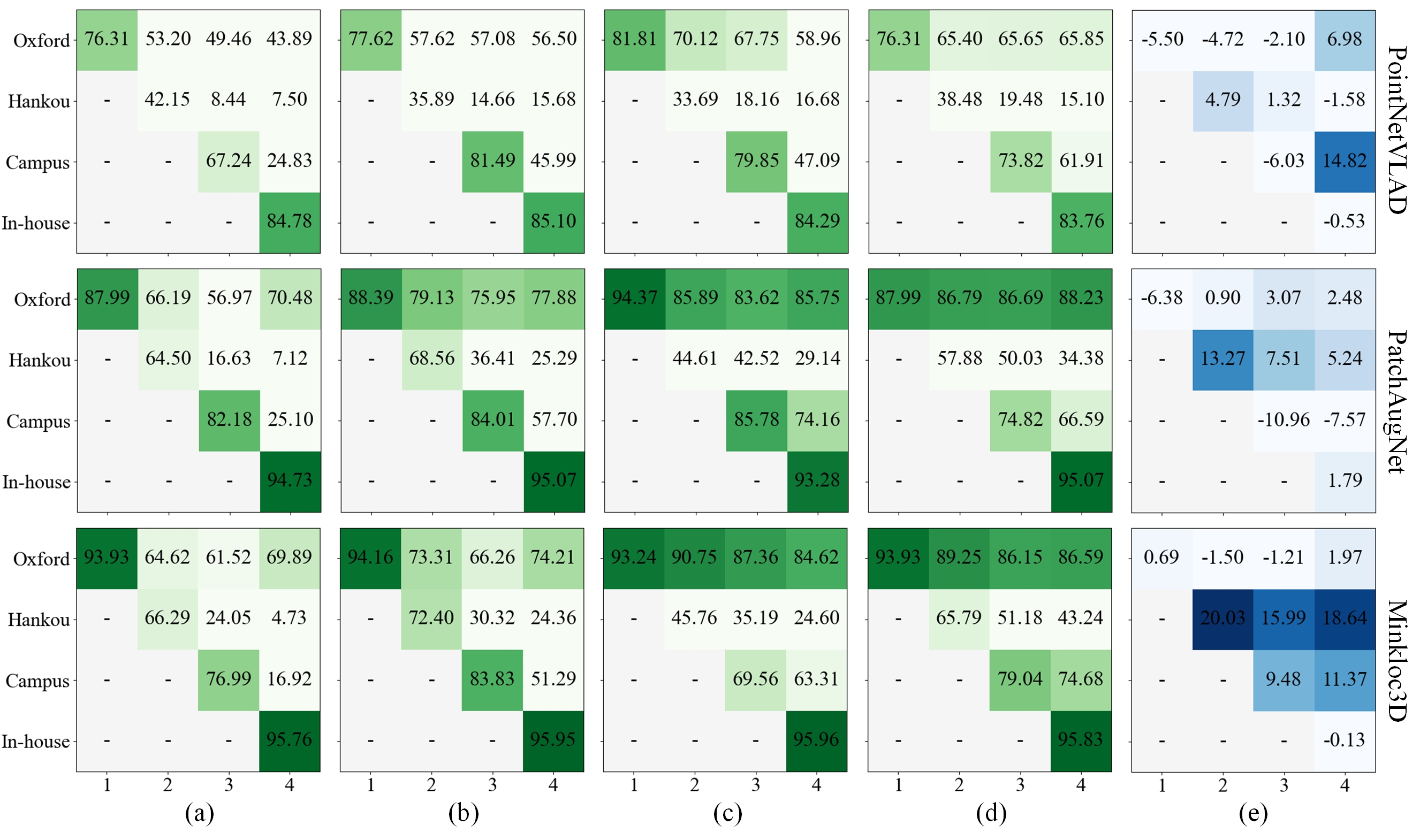}
    \caption{Average Recall@1 on Seq2. (a) FT, (b) InCloud, (c) CCL (SOTA method), (d) LifelongPR (ours), (d) improvements of LifelongPR compared to CCL. Rows 1-3 correspond to different backbone networks.}
    \label{fig:LifelongPR-comparison-res-hete}
\end{figure*}

Table \ref{tab:comparison-res-seq2} and Fig. \ref{fig:LifelongPR-comparison-res-hete} are the quantitative evaluation results of all methods on Seq2. As shown in Table \ref{tab:comparison-res-seq2} and Fig. \ref{fig:LifelongPR-comparison-res-hete}, LifelongPR continues to achieve SOTA performance on the more challenging dataset sequence Seq2. Compared with the SOTA method CCL, it demonstrates significant improvements of 6.50\% in $mIR@1$, 7.96\% in $mR@1$, while reducing $F$ by 8.95\%. Notably, the performance improvements are most pronounced when employing Minkloc3D as the backbone network. With Minkloc3D, LifelongPR attains best CL results, despite not employing contrastive learning like CCL to enhance network generalization and mitigate catastrophic forgetting.

\begin{table}
\centering
\fontsize{7.5}{10}\selectfont
\caption{Continual learning trained on Seq2. The red and green numbers indicate the performance gains of our method over the SOTA approach (CCL).}
\label{tab:comparison-res-seq2}
\begin{tabular}{ccccc}
\hline
\textbf{Backbone} & \textbf{CL Method}  & \textbf{mIR@1(\%)$\uparrow$}                             & \textbf{mR@1(\%)$\uparrow$}                          & \textbf{F(\%)$\downarrow$}                             \\
\hline
\multirow{4}{*}{PointNetVLAD}      & FT                  & 51.49                                      & 40.25                                  & 36.50                                  \\
                  & InCloud             & 56.57                                      & 50.82                                  & 25.61                                  \\
                  & CCL                 & \textbf{60.18}                             & \underline{51.76}                                  & \underline{24.20}                                  \\
                  & \textbf{LifelongPR} & \underline{59.47}\scalebox{0.7}{\textcolor{red}{-0.71}} & \textbf{56.66}\scalebox{0.7}{\textcolor[rgb]{0,0.86,0}{+4.90}} & \textbf{15.25}\scalebox{0.7}{\textcolor[rgb]{0,0.86,0}{-8.95}} \\
\hline
\multirow{4}{*}{PatchAugNet}       & FT                  & 63.65                                      & 49.36                                  & 43.99                                  \\
                  & InCloud             & 72.92                                      & 63.99                                  & 26.70                                  \\
                  & CCL                 & \underline{75.21}                                      & \underline{70.58}                                  & \underline{11.90}                                  \\
                  & \textbf{LifelongPR} & \textbf{75.48}\scalebox{0.7}{\textcolor[rgb]{0,0.86,0}{+0.27}}     & \textbf{71.07}\scalebox{0.7}{\textcolor[rgb]{0,0.86,0}{+0.49}} & \textbf{10.58}\scalebox{0.7}{\textcolor[rgb]{0,0.86,0}{-1.32}} \\
\hline
\multirow{4}{*}{Minkloc3D}         & FT                  & 65.10                                      & 46.83                                  & 48.56                                  \\
                  & InCloud             & 72.15                                      & 61.45                                  & 33.51                                  \\
                  & CCL                 & \underline{73.16}                                      & \underline{67.12}                                  & \underline{12.01}                                  \\
                  & \textbf{LifelongPR} & \textbf{79.67}\scalebox{0.7}{\textcolor[rgb]{0,0.86,0}{+6.50}}     & \textbf{75.08}\scalebox{0.7}{\textcolor[rgb]{0,0.86,0}{+7.96}} & \textbf{11.42}\scalebox{0.7}{\textcolor[rgb]{0,0.86,0}{-0.59}} \\
\hline
\end{tabular}
\end{table}

\begin{figure*}
    \centering
    \includegraphics[width=1\linewidth]{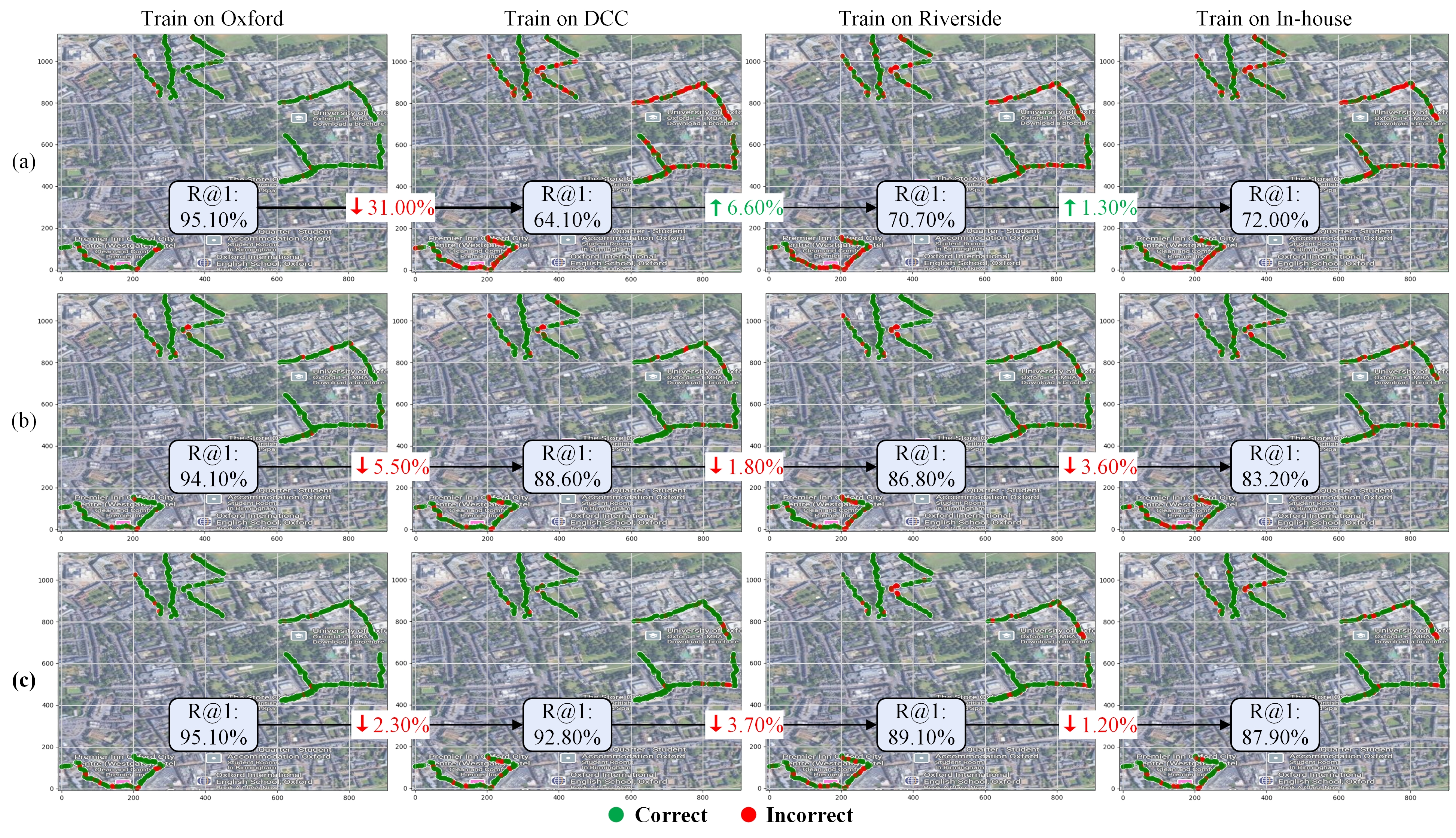}
    \caption{Place recognition results on Oxford after continual learning on Seq1. (a) FT ($\downarrow$23.10\%), (b) CCL ($\downarrow$10.90\%), (d) LifelongPR (ours, $\downarrow$7.20\%). $R@1$ means top-1 recall and only 1000 queries are selected from Oxford for better representation.}
    \label{fig:LifelongPR-comparison-res-dcc-part}
\end{figure*}

Fig. \ref{fig:LifelongPR-comparison-res-dcc-part} presents the place recognition results on Oxford after continual learning on Seq1. As shown in Fig. \ref{fig:LifelongPR-comparison-res-dcc-part} (a), after continual learning on Seq1 with FT, top-1 recall drop from 95.10\% to 72.00\%, declining by 23.10\%. As shown in Fig. \ref{fig:LifelongPR-comparison-res-dcc-part} (b), after continual learning on Seq1 with CCL, top-1 recall drop from 94.10\% to 83.20\%, declining by 10.90\%. As shown in Fig. \ref{fig:LifelongPR-comparison-res-dcc-part} (c), after continual learning on Seq1 with LifelongPR, top-1 recall drop from 95.10\% to 87.90\%, declining by only 7.20\%. Our method LifelongPR achieves the best performance, exhibiting excellent anti-forgetting capabilities.

Our method consistently achieves the best CL performance across different backbone networks and datasets of varying difficulty levels. This superior performance stems from three key factors: 1) Samples in certain datasets exhibit uneven spatial distributions. A representative example is the Oxford dataset, where the 45 collection trajectories within the same area demonstrate significant variations, resulting in some regions being oversampled while others are undersampled. The proposed replay sample selection method, which explicitly considers spatial distribution characteristics, effectively addresses this challenge. 2) The proposed method addresses the significant variations in scene scale and diversity across different datasets by quantitatively evaluating the information quantity of each training set to allocate the replay sample size accordingly, thereby enabling more efficient and rational utilization of the limited replay memory. 3) The prompt module acts as a lightweight memory component. Through the proposed two-stage training strategy, it effectively learns domain-specific knowledge from different datasets, thereby guiding the backbone network to extract features adapted to individual samples. Additionally, we can further enhance the performance of our method by employing contrastive learning, similar to CCL.

\begin{figure}
    \centering
    \includegraphics[width=1\linewidth]{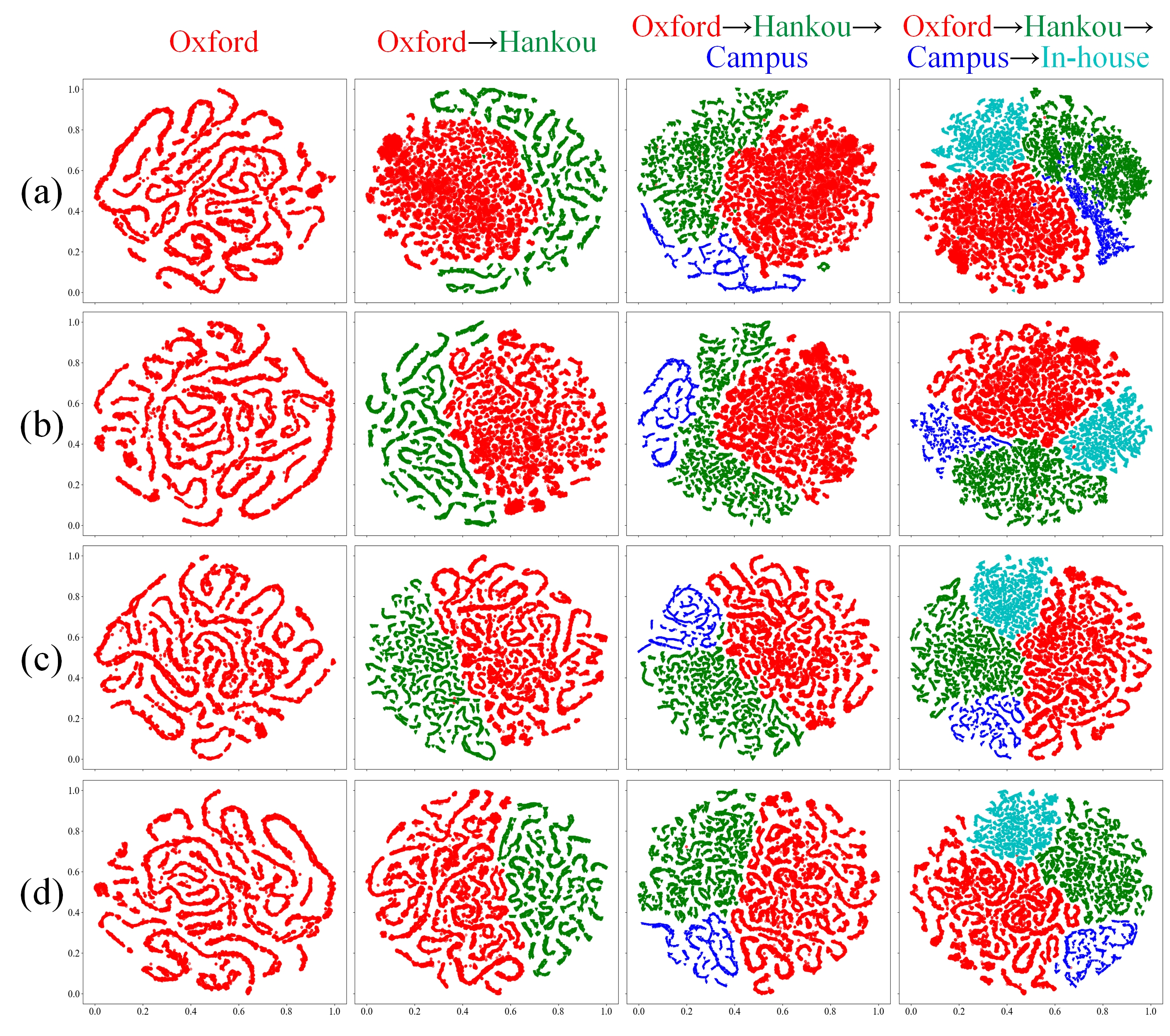}
    \caption{t-SNE visualization of samples in Seq2 at different stages of continuous learning. Rows 1-4 corresponds to different continual learning methods, (a) FT, (b) InCloud, (b) CCL, (d) LifelongPR (ours).}
    \label{fig:LifelongPR-comparison-res-hete-tsne}
\end{figure}

\textbf{t-SNE visualization.}
Fig. \ref{fig:LifelongPR-comparison-res-hete-tsne} is the t-SNE visualization of samples in Seq2 at different stages of CL. As shown in row 1, after training on Oxford, samples of Oxford exhibit a relatively dispersed distribution in feature space, indicating strong discriminability. However, after finishing CL on Seq2, samples of Oxford gradually become more clustered in feature space, showing reduced discriminability. This demonstrates catastrophic forgetting. Rows 2-4 show that sample replay, knowledge distillation, and prompt learning effectively mitigate catastrophic forgetting of the PCPR model. Among these methods, CCL and the proposed method LifelongPR perform the best, maintaining dispersed distributions of samples from each dataset in feature space across different learning stages. It demonstrates that our method has strong anti-forgetting capabilities, consistent with quantitative results in Table \ref{tab:comparison-res-seq2} and Fig. \ref{fig:LifelongPR-comparison-res-hete}.

\begin{figure*}
    \centering
    \includegraphics[width=0.9\linewidth]{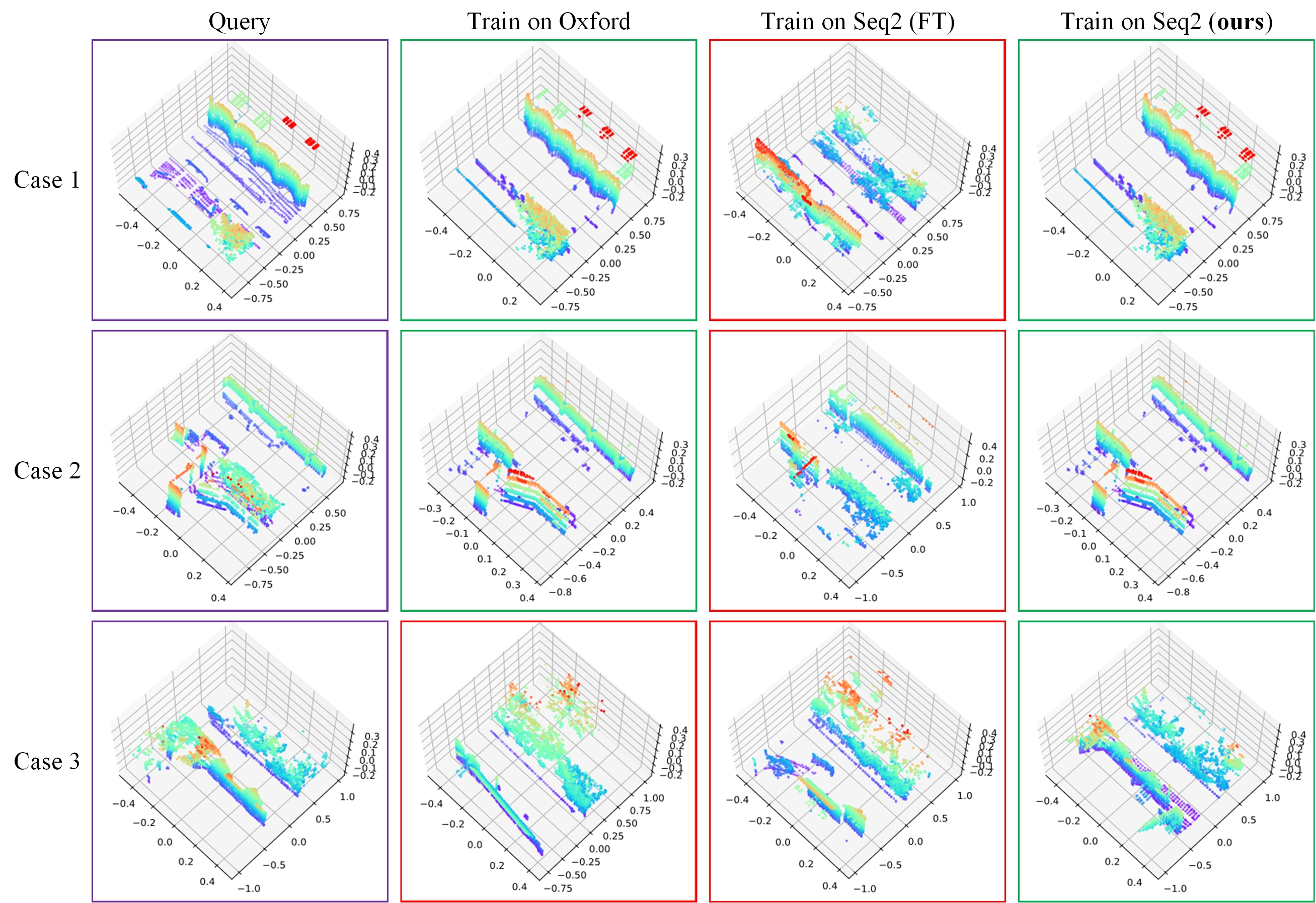}
    \caption{Three success cases of LifelongPR on Seq2. Column 1: query; Column 2: top 1 submaps retrieved by the model trained on Oxford; Columns 3-4: top 1 submaps retrieved by the model trained on Seq2.}
    \label{fig:LifelongPR-success-cases}
\end{figure*}

\textbf{Success cases.}
Fig. \ref{fig:LifelongPR-success-cases} presents three success cases of our method in typical street scenes. In cases 1-2, the model trained on Oxford successfully retrieves the correct top-1 submaps. However, retrieval fails after CL using FT, whereas it succeeds when LifelongPR is used. In case 3, the model trained on Oxford fails to retrieve the correct top-1 submap, and retrieval still fails after CL using FT, but succeeds when LifelongPR is used. Analysis of specific cases reveals that the original PCPR model has strong capabilities in extracting scene detail information. After CL using FT, the model's local feature extraction ability significantly declines, leading to retrieval failures. In contrast, our method effectively mitigates the decline in feature extraction capability through more reasonable replay sample selection and a prompt learning-based CL framework.

\section{Analysis and discussion}\label{sec_analysis_discussion}
This section analyzes the key modules, critical parameters, and limitations of the proposed method on Seq2, while also providing an outlook on future work.

\subsection{Effectiveness of replay sample selection method}\label{subsec_sample_selection}
\begin{figure}[h]
    \centering
    \includegraphics[width=1\linewidth]{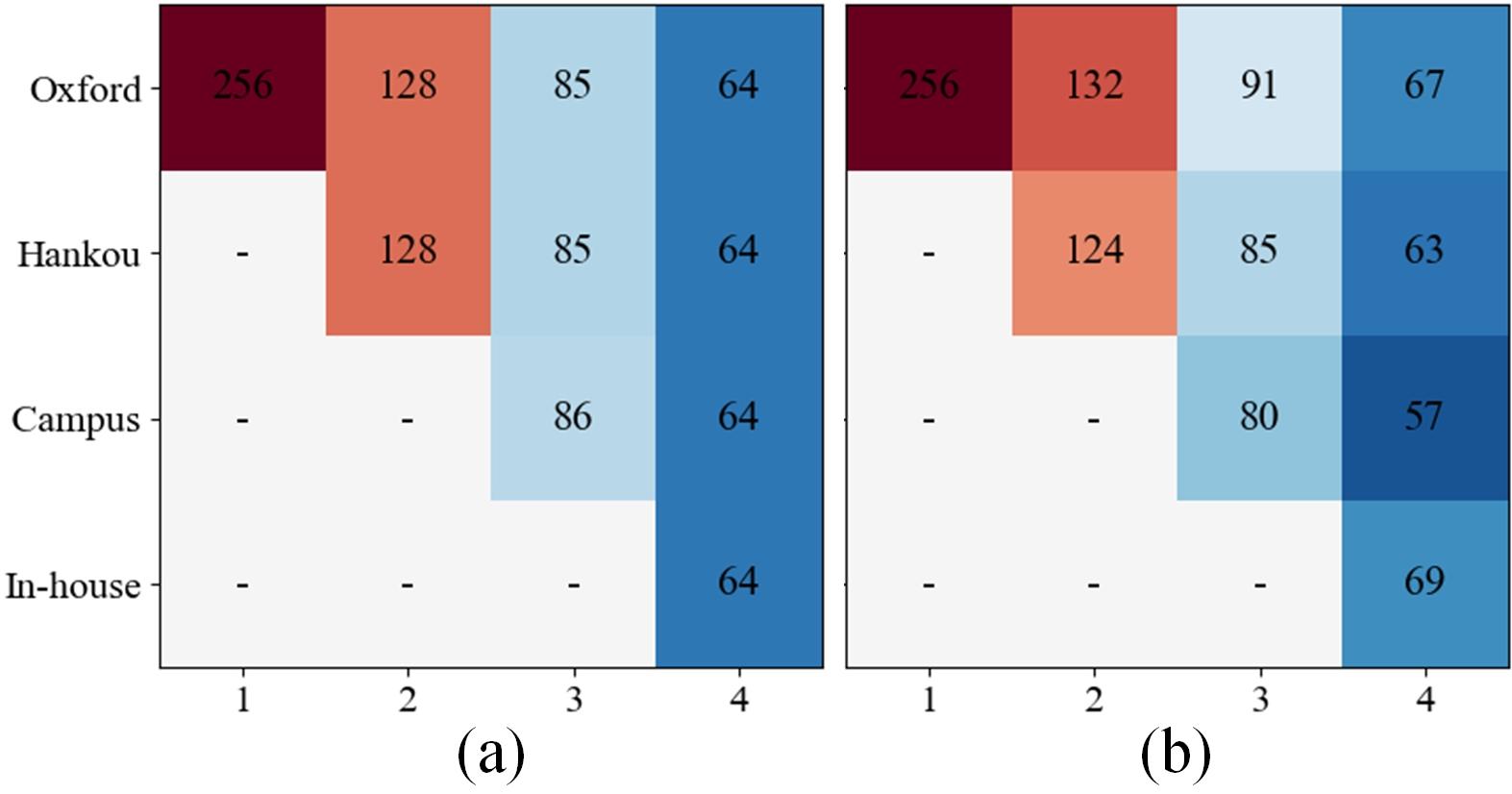}
    \caption{Replay sample size allocation results. (a) not use information quantity, (b) use information quantity (ours).}
    \label{fig:LifelongPR-ablation-information-quantity}
\end{figure}

Table \ref{tab:ablation-key-module} presents the ablation experiment results of the core modules. (a) utilizes random sampling for replay sample selection, yielding poor CL performance. In contrast, (b), (c), and (d) demonstrate significant performance improvements, with $mIR@1$ increasing by 0.31\%–0.69\% and $mR@1$ by 0.95\%–1.36\%, while $F$ decreases by 1.38\%-4.11\%. This performance improvement stems from our sophisticated replay sample selection method, which considers sample distributions in both Euclidean and feature spaces, thereby identifying and retaining more informative samples for replaying. On the basis of (d), (e) further enhances performance by incorporating information quantity differences across datasets when allocating replay sample size, achieving additional improvements of 0.89\% in $mIR@1$ and 0.60\% in $mR@1$. Fig. \ref{fig:LifelongPR-ablation-information-quantity} (b) shows the replay sample size allocation results. Compared to uniformly allocating replay sample sizes across all training sets, our method allocates a greater number of replay samples to more informative training sets, such as Oxford and In-house, enabling more effective utilization of limited replay sample budget.

\begin{table*}
\centering
\caption{Ablation experiment results of key modules including replay sample selection and the prompt module. Random: randomly sampling; Greedy: sampling based on greedy algorithm; InfoQ: allocating sample size based on information quantity.}
\label{tab:ablation-key-module}
\begin{tabular}{ccccccc}
\toprule
\textbf{ID} & \textbf{Select} & \textbf{Forget} & \textbf{Prompt} & \textbf{mIR@1(\%)$\uparrow$} & \textbf{mR@1(\%)$\uparrow$} & \textbf{F(\%)$\downarrow$}  \\ 
\midrule
(a)         & Random          & Random          & \ding{55}              & 77.43          & 70.84         & 18.55       \\
(b)         & Random          & Greedy          & \ding{55}              & 78.12          & 71.93         & 17.17       \\
(c)         & Greedy          & Random          & \ding{55}              & 77.74          & 71.79         & 15.29       \\
(d)         & Greedy          & Greedy          & \ding{55}              & 77.65          & 72.20         & 14.44       \\
(e)         & Greedy-InfoQ    & Greedy-InfoQ    & \ding{55}              & 78.54          & 72.80         & 15.59       \\
(f)         & Greedy-InfoQ    & Greedy-InfoQ    & \ding{51}              & 79.67          & 75.08         & 11.42       \\
\bottomrule
\end{tabular}
\end{table*}

\subsection{Effectiveness of prompt module}\label{subsec_prompt_learning}
As shown in Table \ref{tab:ablation-key-module}, on the basis of (e), (f) further enhances performance by employing the prompt module to extract domain-specific knowledge from each dataset and guiding the backbone network to to extract features adapted to samples. This yields additional improvements of 1.13\% in $mIR@1$ and 2.28\% in $mR@1$, while reducing $F$ by 4.17\%. During CL, the prompt module effectively serves as an external memory unit. By combining a two-stage training strategy with replay samples, it successfully captures distinctive domain-specific knowledge from each dataset. At inference time, the encoded representations generated by the prompt module are fed as auxiliary inputs to the backbone network, directing it to extract sample-adaptive global features.

\subsection{Effectiveness of two-stage training strategy}\label{subsec_training_strategy}
\begin{figure*}[h]
    \centering
    \includegraphics[width=1\linewidth]{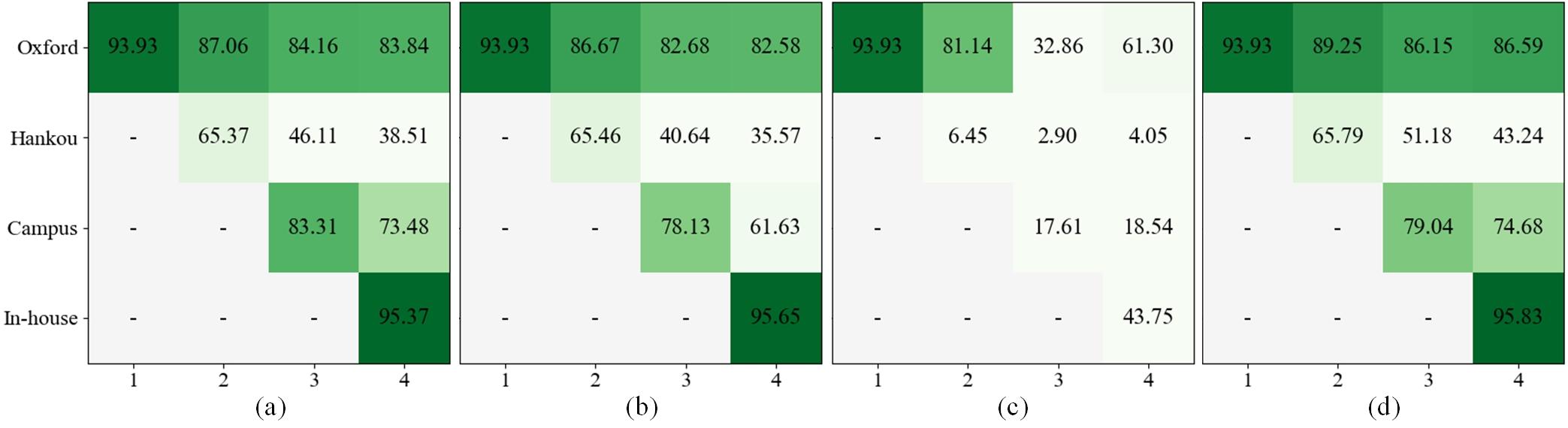}
    \caption{Average Recall@1 results of LifelongPR with different training strategies. (a) no prompt, (b) one-stage, (c) prompt only, (d) two-stage (ours).}
    \label{fig:LifelongPR-ablation-training-strategy}
\end{figure*}

Table \ref{tab:ablation-training-strategy} and Fig. \ref{fig:LifelongPR-ablation-training-strategy} present CL results under different training strategies, where our two-stage strategy achieves best performance while training only the prompt module performs worst. Comparative analysis of Fig. \ref{fig:LifelongPR-ablation-training-strategy} (a)-(c) demonstrates that the naive addition of a prompt module fails to improve the model's anti-forgetting capabilities, regardless of one-stage joint training or prompt-only training. As shown in Table \ref{tab:ablation-training-strategy} and Fig. \ref{fig:LifelongPR-ablation-training-strategy} (d), the two-stage training strategy shows significant improvements over the one-stage training strategy, with gains of 3.17\% in $mIR@1$, 6.22\% in $mR@1$, and a 7.83\% reduction in $F$. In stage 1, the prompt module is trained to extract domain-specific knowledge from the current dataset, while in stage 2, the backbone network is fine-tuned to adapt to the target domain of the training set. The experimental results demonstrate the rationality and effectiveness of this two-stage training strategy.
\begin{table}
\centering
\caption{Continual learning results of LifelongPR with different training strategies.}
\label{tab:ablation-training-strategy}
\begin{tabular}{cccc} 
\hline
\textbf{Train Strategy} & \textbf{mIR@1(\%)$\uparrow$} & \textbf{mR@1(\%)$\uparrow$}  & \textbf{F(\%)$\downarrow$}      \\ 
\hline
One-stage               & \underline{76.50}          & \underline{68.86}          & 19.25           \\
Prompt only             & 46.86          & 31.91          & \textbf{11.37}  \\
\textbf{Two-stage}      & \textbf{79.67} & \textbf{75.08} & \underline{11.42}           \\
\hline
\end{tabular}
\end{table}

\subsection{Parameter analysis}\label{subsec_param_analysis}
Table \ref{tab:param-analysis} presents the CL results of LifelongPR with different temperature parameters. The temperature parameter $\tau$ serves as a key hyperparameter in replay sample selection, where larger values lead to more uniform allocation of replay samples. Experimental results demonstrate that our method achieves optimal performance when $\tau$ is set to 4.0, which is therefore recommended as the default parameter setting.

\begin{table}
\centering
\caption{Continual learning results of LifelongPR with different temperature thresholds $\tau$. Note: The prompt module is not employed.}
\label{tab:param-analysis}
\begin{tabular}{cccc} 
\hline
\textbf{Temperature ($\tau$)} & \textbf{mIR@1(\%)$\uparrow$} & \textbf{mR@1(\%)$\uparrow$}  & \textbf{F(\%)$\downarrow$}      \\ 
\hline
0.5  & 77.07          & 69.78          & 16.88           \\
1.0  & 77.73          & \underline{71.91}          & 16.01           \\
2.0  & 77.37          & 70.06          & 17.91           \\
\textbf{4.0}& \textbf{78.54} & \textbf{72.80} & \underline{15.59}  \\
8.0  & 77.03          & 71.86          & \textbf{14.93}           \\
16.0 & \underline{77.87}          & 71.17          & 17.70           \\
\hline
\end{tabular}
\end{table}

\subsection{Deficiencies and future work}\label{subsec_deficiency_future_work}
\begin{figure*}
    \centering
    \includegraphics[width=0.9\linewidth]{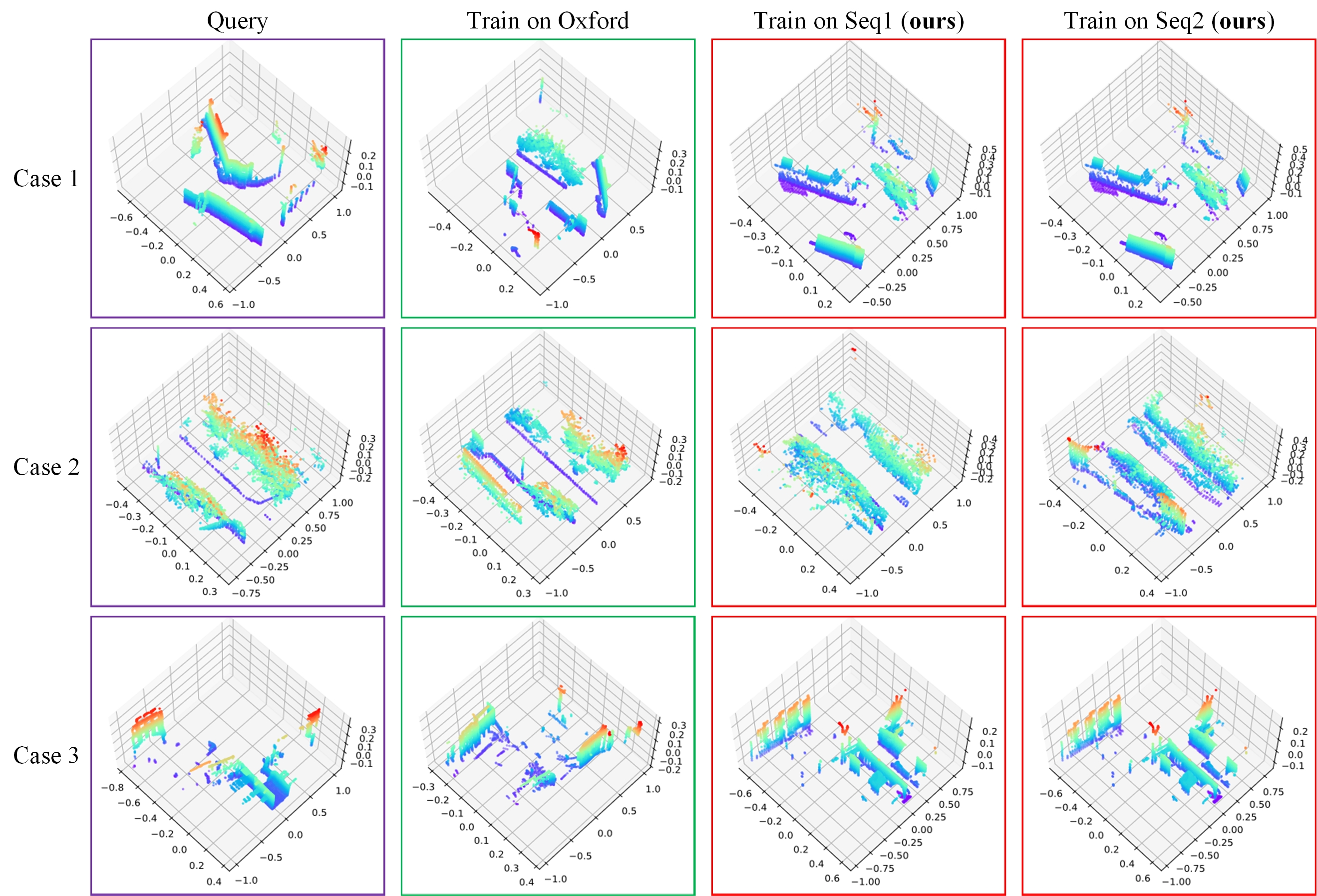}
    \caption{Three bad cases of LifelongPR trained on Seq1 and Seq2. Column 1: query; Column 2: top 1 submaps retrieved by the model trained on Oxford; Column 3: top 1 submaps retrieved by the model trained on Seq1; Column 4: top 1 submaps retrieved by the model trained on Seq2.}
    \label{fig:LifelongPR-bad-cases}
\end{figure*}

While our method demonstrates promising performance, several limitations remain. Fig. \ref{fig:LifelongPR-bad-cases} presents three representative failure cases. Case 1 involves a T-junction, while Case 3 exhibits insufficient scene characteristics in the query submap, where both the retrieved top 1 submaps and the query submap show high similarity. Case 2 represents a typical street scene with trees lining the roadsides, lacking distinctive geographical features. Among these cases, Cases 1 and 2 could potentially be addressed by enhancing network generalization and anti-forgetting capabilities. However, Case 3 involves repetitive scenes that cannot be resolved through single-shot retrieval. Furthermore, the proposed prompt module also exhibits catastrophic forgetting, and the CL approach built upon it necessitates that the backbone network must be initially trained to ensure robust feature extraction capability and strong generalizability. In the future, we plan to incorporate theories and approaches of domain generalization \citep{zhou2022domain} and contrastive learning \citep{liu2021self} to enhance the backbone network's feature extraction capability and generalizability, thereby fundamentally improving the anti-forgetting capability of PCPR models. Furthermore, we will research on online CL for PCPR, enabling the model to evolve incrementally during runtime operation \citep{aljundi2019gradient}.

\section{Conclusion}\label{sec_conclusion}
To address catastrophic forgetting of PCPR models, this study proposes a novel CL method tailored for PCPR, effectively extracting and fusing knowledge learned across sequential point cloud data. First, a replay sample selection method is proposed, dynamically allocating sample sizes to each training set based on information quantity and selecting replay samples based on spatial distribution. Second,  a new CL framework composed of a prompt module and a two-stage training strategy is proposed, and the domain-specific knowledge captured from each training set by the prompt module is used for guiding the backbone network to extract sample-adaptive features. We validate the effectiveness of the proposed method on large-scale public and self-collected datasets, demonstrating significant improvements over SOTA methods: 6.50\% increase in $mIR@1$, 7.96\% increase in $mR@1$, and 8.95\% reduction in $F$. Furthermore, we conduct comprehensive experimental analyses for key components, critical parameters, and limitations of the proposed method. In the future, we plan to leverage the theories of domain generalization and contrastive learning to fundamentally enhance the model's anti-forgetting capability.

\bibliography{LifelongPR} 

\begin{thebibliography}{61}
\providecommand{\natexlab}[1]{#1}
\providecommand{\url}[1]{#1}
\csname url@samestyle\endcsname
\providecommand{\newblock}{\relax}
\providecommand{\bibinfo}[2]{#2}
\providecommand{\BIBentrySTDinterwordspacing}{\spaceskip=0pt\relax}
\providecommand{\BIBentryALTinterwordstretchfactor}{4}
\providecommand{\BIBentryALTinterwordspacing}{\spaceskip=\fontdimen2\font plus
\BIBentryALTinterwordstretchfactor\fontdimen3\font minus \fontdimen4\font\relax}
\providecommand{\BIBforeignlanguage}[2]{{%
\expandafter\ifx\csname l@#1\endcsname\relax
\typeout{** WARNING: IEEEtranN.bst: No hyphenation pattern has been}%
\typeout{** loaded for the language `#1'. Using the pattern for}%
\typeout{** the default language instead.}%
\else
\language=\csname l@#1\endcsname
\fi
#2}}
\providecommand{\BIBdecl}{\relax}
\BIBdecl

\bibitem[Zhou et~al.(2025)Zhou, Long, Xie, Wang, Zhang, Li, Wang, Chen, and Dong]{zhou2025whu}
J.~Zhou, C.~Long, Y.~Xie, J.~Wang, C.~Zhang, B.~Li, H.~Wang, Z.~Chen, and Z.~Dong, ``Whu-synthetic: A synthetic perception dataset for 3d multi-task model research,'' \emph{IEEE Transactions on Geoscience and Remote Sensing}, 2025.

\bibitem[Li et~al.(2025{\natexlab{a}})Li, Li, Dong, Wang, and Yang]{li2025saliencyi2ploc}
Y.~Li, J.~Li, Z.~Dong, Y.~Wang, and B.~Yang, ``Saliencyi2ploc: saliency-guided image-point cloud localization using contrastive learning,'' \emph{Information Fusion}, vol. 118, p. 103015, 2025.

\bibitem[Yurtsever et~al.(2020)Yurtsever, Lambert, Carballo, and Takeda]{yurtsever2020survey}
E.~Yurtsever, J.~Lambert, A.~Carballo, and K.~Takeda, ``A survey of autonomous driving: Common practices and emerging technologies,'' \emph{IEEE access}, vol.~8, pp. 58\,443--58\,469, 2020.

\bibitem[Li et~al.(2023{\natexlab{a}})Li, Wu, Yang, Zou, Yang, Zhao, and Dong]{li2023whu}
J.~Li, W.~Wu, B.~Yang, X.~Zou, Y.~Yang, X.~Zhao, and Z.~Dong, ``Whu-helmet: A helmet-based multisensor slam dataset for the evaluation of real-time 3-d mapping in large-scale gnss-denied environments,'' \emph{IEEE Transactions on Geoscience and Remote Sensing}, vol.~61, pp. 1--16, 2023.

\bibitem[Ounoughi and Yahia(2023)]{ounoughi2023data}
C.~Ounoughi and S.~B. Yahia, ``Data fusion for its: A systematic literature review,'' \emph{Information Fusion}, vol.~89, pp. 267--291, 2023.

\bibitem[Li et~al.(2024)Li, Yuan, Cao, Nguyen, Cao, and Xie]{li2024hcto}
J.~Li, S.~Yuan, M.~Cao, T.-M. Nguyen, K.~Cao, and L.~Xie, ``Hcto: Optimality-aware lidar inertial odometry with hybrid continuous time optimization for compact wearable mapping system,'' \emph{ISPRS Journal of Photogrammetry and Remote Sensing}, vol. 211, pp. 228--243, 2024.

\bibitem[Xu et~al.(2025)Xu, Chen, Yang, Wu, Sun, Wang, Li, and Zou]{xu2025atcm}
Y.~Xu, C.~Chen, B.~Yang, W.~Wu, S.~Sun, Z.~Wang, L.~Li, and Q.~Zou, ``Atcm: Aerial-terrestrial lidar-based collaborative simultaneous localization and mapping,'' \emph{IEEE Transactions on Geoscience and Remote Sensing}, 2025.

\bibitem[Buehrer et~al.(2018)Buehrer, Wymeersch, and Vaghefi]{buehrer2018collaborative}
R.~M. Buehrer, H.~Wymeersch, and R.~M. Vaghefi, ``Collaborative sensor network localization: Algorithms and practical issues,'' \emph{Proceedings of the IEEE}, vol. 106, no.~6, pp. 1089--1114, 2018.

\bibitem[Carmigniani et~al.(2011)Carmigniani, Furht, Anisetti, Ceravolo, Damiani, and Ivkovic]{carmigniani2011augmented}
J.~Carmigniani, B.~Furht, M.~Anisetti, P.~Ceravolo, E.~Damiani, and M.~Ivkovic, ``Augmented reality technologies, systems and applications,'' \emph{Multimedia tools and applications}, vol.~51, pp. 341--377, 2011.

\bibitem[Li et~al.(2025{\natexlab{b}})Li, Xu, Liu, Cao, Yuan, and Xie]{li2025ua}
J.~Li, X.~Xu, J.~Liu, K.~Cao, S.~Yuan, and L.~Xie, ``Ua-mpc: Uncertainty-aware model predictive control for motorized lidar odometry,'' \emph{IEEE Robotics and Automation Letters}, 2025.

\bibitem[Uy and Lee(2018)]{uy2018pointnetvlad}
M.~A. Uy and G.~H. Lee, ``Pointnetvlad: Deep point cloud based retrieval for large-scale place recognition,'' in \emph{Proceedings of the IEEE conference on computer vision and pattern recognition}, 2018, pp. 4470--4479.

\bibitem[Komorowski(2021)]{komorowski2021minkloc3d}
J.~Komorowski, ``Minkloc3d: Point cloud based large-scale place recognition,'' in \emph{Proceedings of the IEEE/CVF Winter Conference on Applications of Computer Vision}, 2021, pp. 1790--1799.

\bibitem[Zou et~al.(2023)Zou, Li, Wang, Liang, Wu, Wang, Yang, and Dong]{zou2023patchaugnet}
X.~Zou, J.~Li, Y.~Wang, F.~Liang, W.~Wu, H.~Wang, B.~Yang, and Z.~Dong, ``Patchaugnet: Patch feature augmentation-based heterogeneous point cloud place recognition in large-scale street scenes,'' \emph{ISPRS Journal of Photogrammetry and Remote Sensing}, vol. 206, pp. 273--292, 2023.

\bibitem[Jung et~al.(2024)Jung, Yang, Lee, Gil, Kim, and Kim]{jung2024helipr}
M.~Jung, W.~Yang, D.~Lee, H.~Gil, G.~Kim, and A.~Kim, ``Helipr: Heterogeneous lidar dataset for inter-lidar place recognition under spatiotemporal variations,'' \emph{The International Journal of Robotics Research}, vol.~43, no.~12, pp. 1867--1883, 2024.

\bibitem[Lesort et~al.(2020)Lesort, Lomonaco, Stoian, Maltoni, Filliat, and D{\'\i}az-Rodr{\'\i}guez]{lesort2020continual}
T.~Lesort, V.~Lomonaco, A.~Stoian, D.~Maltoni, D.~Filliat, and N.~D{\'\i}az-Rodr{\'\i}guez, ``Continual learning for robotics: Definition, framework, learning strategies, opportunities and challenges,'' \emph{Information fusion}, vol.~58, pp. 52--68, 2020.

\bibitem[Aljundi et~al.(2017)Aljundi, Chakravarty, and Tuytelaars]{aljundi2017expert}
R.~Aljundi, P.~Chakravarty, and T.~Tuytelaars, ``Expert gate: Lifelong learning with a network of experts,'' in \emph{Proceedings of the IEEE conference on computer vision and pattern recognition}, 2017, pp. 3366--3375.

\bibitem[Serra et~al.(2018)Serra, Suris, Miron, and Karatzoglou]{serra2018overcoming}
J.~Serra, D.~Suris, M.~Miron, and A.~Karatzoglou, ``Overcoming catastrophic forgetting with hard attention to the task,'' in \emph{International conference on machine learning}, 2018, pp. 4548--4557.

\bibitem[Zheng et~al.(2025)Zheng, Zhang, Tian, and Du]{zheng2025enhancing}
Y.~Zheng, X.~Zhang, Z.~Tian, and S.~Du, ``Enhancing few-shot lifelong learning through fusion of cross-domain knowledge,'' \emph{Information Fusion}, vol. 115, p. 102730, 2025.

\bibitem[Knights et~al.(2022)Knights, Moghadam, Ramezani, Sridharan, and Fookes]{knights2022incloud}
J.~Knights, P.~Moghadam, M.~Ramezani, S.~Sridharan, and C.~Fookes, ``Incloud: Incremental learning for point cloud place recognition,'' in \emph{2022 IEEE/RSJ International Conference on Intelligent Robots and Systems (IROS)}, 2022, pp. 8559--8566.

\bibitem[Cui and Chen(2023)]{cui2023ccl}
J.~Cui and X.~Chen, ``Ccl: Continual contrastive learning for lidar place recognition,'' \emph{IEEE Robotics and Automation Letters}, vol.~8, no.~8, pp. 4433--4440, 2023.

\bibitem[Rebuffi et~al.(2017)Rebuffi, Kolesnikov, Sperl, and Lampert]{rebuffi2017icarl}
S.-A. Rebuffi, A.~Kolesnikov, G.~Sperl, and C.~H. Lampert, ``icarl: Incremental classifier and representation learning,'' in \emph{Proceedings of the IEEE conference on Computer Vision and Pattern Recognition}, 2017, pp. 2001--2010.

\bibitem[Li and Hoiem(2017)]{li2017learning}
Z.~Li and D.~Hoiem, ``Learning without forgetting,'' \emph{IEEE transactions on pattern analysis and machine intelligence}, vol.~40, no.~12, pp. 2935--2947, 2017.

\bibitem[Schaupp et~al.(2019)Schaupp, B{\"u}rki, Dub{\'e}, Siegwart, and Cadena]{schaupp2019oreos}
L.~Schaupp, M.~B{\"u}rki, R.~Dub{\'e}, R.~Siegwart, and C.~Cadena, ``Oreos: Oriented recognition of 3d point clouds in outdoor scenarios,'' in \emph{2019 IEEE/RSJ International Conference on Intelligent Robots and Systems (IROS)}, 2019, pp. 3255--3261.

\bibitem[Kim et~al.(2019)Kim, Park, and Kim]{kim2019sci-loc}
G.~Kim, B.~Park, and A.~Kim, ``1-day learning, 1-year localization: Long-term lidar localization using scan context image,'' \emph{IEEE Robotics and Automation Letters}, vol.~4, no.~2, pp. 1948--1955, 2019.

\bibitem[Kim and Kim(2018)]{kim2018scan}
G.~Kim and A.~Kim, ``Scan context: Egocentric spatial descriptor for place recognition within 3d point cloud map,'' in \emph{2018 IEEE/RSJ International Conference on Intelligent Robots and Systems (IROS)}, 2018, pp. 4802--4809.

\bibitem[Chen et~al.(2021)Chen, L{\"a}be, Milioto, R{\"o}hling, Vysotska, Haag, Behley, and Stachniss]{chen2021overlapnet}
X.~Chen, T.~L{\"a}be, A.~Milioto, T.~R{\"o}hling, O.~Vysotska, A.~Haag, J.~Behley, and C.~Stachniss, ``Overlapnet: Loop closing for lidar-based slam,'' \emph{arXiv preprint arXiv:2105.11344}, 2021.

\bibitem[Komorowski(2022)]{komorowski2022improving}
J.~Komorowski, ``Improving point cloud based place recognition with ranking-based loss and large batch training,'' in \emph{2022 26th international conference on pattern recognition (ICPR)}, 2022, pp. 3699--3705.

\bibitem[Wang et~al.(2020)Wang, Wu, Zhu, Li, Zuo, and Hu]{wang2020eca}
Q.~Wang, B.~Wu, P.~Zhu, P.~Li, W.~Zuo, and Q.~Hu, ``Eca-net: Efficient channel attention for deep convolutional neural networks,'' in \emph{Proceedings of the IEEE/CVF conference on computer vision and pattern recognition}, 2020, pp. 11\,534--11\,542.

\bibitem[Fan et~al.(2022)Fan, Song, Liu, Lu, He, and Du]{fan2022svt}
Z.~Fan, Z.~Song, H.~Liu, Z.~Lu, J.~He, and X.~Du, ``Svt-net: Super light-weight sparse voxel transformer for large scale place recognition,'' in \emph{Proceedings of the AAAI conference on artificial intelligence}, vol.~36, no.~1, 2022, pp. 551--560.

\bibitem[Qi et~al.(2017)Qi, Su, Mo, and Guibas]{qi2017pointnet}
C.~R. Qi, H.~Su, K.~Mo, and L.~J. Guibas, ``Pointnet: Deep learning on point sets for 3d classification and segmentation,'' in \emph{Proceedings of the IEEE conference on computer vision and pattern recognition}, 2017, pp. 652--660.

\bibitem[Arandjelovic et~al.(2016)Arandjelovic, Gronat, Torii, Pajdla, and Sivic]{arandjelovic2016netvlad}
R.~Arandjelovic, P.~Gronat, A.~Torii, T.~Pajdla, and J.~Sivic, ``Netvlad: Cnn architecture for weakly supervised place recognition,'' in \emph{Proceedings of the IEEE conference on computer vision and pattern recognition}, 2016, pp. 5297--5307.

\bibitem[Hui et~al.(2021)Hui, Yang, Cheng, Xie, and Yang]{hui2021pyramid}
L.~Hui, H.~Yang, M.~Cheng, J.~Xie, and J.~Yang, ``Pyramid point cloud transformer for large-scale place recognition,'' in \emph{Proceedings of the IEEE/CVF International Conference on Computer Vision}, 2021, pp. 6098--6107.

\bibitem[Zhou et~al.(2021)Zhou, Zhao, Adolfsson, Su, Gao, Duckett, and Sun]{zhou2021ndt}
Z.~Zhou, C.~Zhao, D.~Adolfsson, S.~Su, Y.~Gao, T.~Duckett, and L.~Sun, ``Ndt-transformer: Large-scale 3d point cloud localisation using the normal distribution transform representation,'' in \emph{2021 IEEE International Conference on Robotics and Automation (ICRA)}, 2021, pp. 5654--5660.

\bibitem[Xie et~al.(2024)Xie, Wang, Wang, Liang, Zhang, Dong, and Yang]{xie2024look}
Y.~Xie, B.~Wang, H.~Wang, F.~Liang, W.~Zhang, Z.~Dong, and B.~Yang, ``Look at the whole scene: General point cloud place recognition by classification proxy,'' \emph{ISPRS Journal of Photogrammetry and Remote Sensing}, vol. 215, pp. 15--30, 2024.

\bibitem[Wang et~al.(2024)Wang, Zhang, Su, and Zhu]{wang2024comprehensive}
L.~Wang, X.~Zhang, H.~Su, and J.~Zhu, ``A comprehensive survey of continual learning: theory, method and application,'' \emph{IEEE Transactions on Pattern Analysis and Machine Intelligence}, vol.~46, no.~8, pp. 5362--5383, 2024.

\bibitem[De~Lange et~al.(2021)De~Lange, Aljundi, Masana, Parisot, Jia, Leonardis, Slabaugh, and Tuytelaars]{de2021continual}
M.~De~Lange, R.~Aljundi, M.~Masana, S.~Parisot, X.~Jia, A.~Leonardis, G.~Slabaugh, and T.~Tuytelaars, ``A continual learning survey: Defying forgetting in classification tasks,'' \emph{IEEE transactions on pattern analysis and machine intelligence}, vol.~44, no.~7, pp. 3366--3385, 2021.

\bibitem[Wu et~al.(2019)Wu, Chen, Wang, Ye, Liu, Guo, and Fu]{wu2019large}
Y.~Wu, Y.~Chen, L.~Wang, Y.~Ye, Z.~Liu, Y.~Guo, and Y.~Fu, ``Large scale incremental learning,'' in \emph{Proceedings of the IEEE/CVF conference on computer vision and pattern recognition}, 2019, pp. 374--382.

\bibitem[Shin et~al.(2017)Shin, Lee, Kim, and Kim]{shin2017continual}
H.~Shin, J.~K. Lee, J.~Kim, and J.~Kim, ``Continual learning with deep generative replay,'' \emph{Advances in neural information processing systems}, vol.~30, 2017.

\bibitem[Wu et~al.(2018)Wu, Herranz, Liu, Van De~Weijer, Raducanu, et~al.]{wu2018memory}
C.~Wu, L.~Herranz, X.~Liu, J.~Van De~Weijer, B.~Raducanu \emph{et~al.}, ``Memory replay gans: Learning to generate new categories without forgetting,'' \emph{Advances in neural information processing systems}, vol.~31, 2018.

\bibitem[MacQueen(1967)]{macqueen1967some}
J.~MacQueen, ``Some methods for classification and analysis of multivariate observations,'' in \emph{Proceedings of the Fifth Berkeley Symposium on Mathematical Statistics and Probability, Volume 1: Statistics}, vol.~5, 1967, pp. 281--298.

\bibitem[Killamsetty et~al.(2021)Killamsetty, Durga, Ramakrishnan, De, and Iyer]{killamsetty2021grad}
K.~Killamsetty, S.~Durga, G.~Ramakrishnan, A.~De, and R.~Iyer, ``Grad-match: Gradient matching based data subset selection for efficient deep model training,'' in \emph{International Conference on Machine Learning}, 2021, pp. 5464--5474.

\bibitem[Kothawade et~al.(2022)Kothawade, Kaushal, Ramakrishnan, Bilmes, and Iyer]{kothawade2022prism}
S.~Kothawade, V.~Kaushal, G.~Ramakrishnan, J.~Bilmes, and R.~Iyer, ``Prism: A rich class of parameterized submodular information measures for guided data subset selection,'' in \emph{Proceedings of the AAAI Conference on Artificial Intelligence}, vol.~36, no.~9, 2022, pp. 10\,238--10\,246.

\bibitem[Kirkpatrick et~al.(2017)Kirkpatrick, Pascanu, Rabinowitz, Veness, Desjardins, Rusu, Milan, Quan, Ramalho, Grabska-Barwinska, et~al.]{kirkpatrick2017overcoming}
J.~Kirkpatrick, R.~Pascanu, N.~Rabinowitz, J.~Veness, G.~Desjardins, A.~A. Rusu, K.~Milan, J.~Quan, T.~Ramalho, A.~Grabska-Barwinska \emph{et~al.}, ``Overcoming catastrophic forgetting in neural networks,'' \emph{Proceedings of the national academy of sciences}, vol. 114, no.~13, pp. 3521--3526, 2017.

\bibitem[Mallya and Lazebnik(2018)]{mallya2018packnet}
A.~Mallya and S.~Lazebnik, ``Packnet: Adding multiple tasks to a single network by iterative pruning,'' in \emph{Proceedings of the IEEE conference on Computer Vision and Pattern Recognition}, 2018, pp. 7765--7773.

\bibitem[Fernando et~al.(2017)Fernando, Banarse, Blundell, Zwols, Ha, Rusu, Pritzel, and Wierstra]{fernando2017pathnet}
C.~Fernando, D.~Banarse, C.~Blundell, Y.~Zwols, D.~Ha, A.~A. Rusu, A.~Pritzel, and D.~Wierstra, ``Pathnet: Evolution channels gradient descent in super neural networks,'' \emph{arXiv preprint arXiv:1701.08734}, 2017.

\bibitem[Yin et~al.(2023)Yin, Abuduweili, Zhao, Xu, Liu, and Scherer]{yin2023bioslam}
P.~Yin, A.~Abuduweili, S.~Zhao, L.~Xu, C.~Liu, and S.~Scherer, ``Bioslam: A bioinspired lifelong memory system for general place recognition,'' \emph{IEEE Transactions on Robotics}, vol.~39, no.~6, pp. 4855--4874, 2023.

\bibitem[Liu et~al.(2023)Liu, Yuan, Fu, Jiang, Hayashi, and Neubig]{liu2023pre}
P.~Liu, W.~Yuan, J.~Fu, Z.~Jiang, H.~Hayashi, and G.~Neubig, ``Pre-train, prompt, and predict: A systematic survey of prompting methods in natural language processing,'' \emph{ACM Computing Surveys}, vol.~55, no.~9, pp. 1--35, 2023.

\bibitem[Houlsby et~al.(2019)Houlsby, Giurgiu, Jastrzebski, Morrone, De~Laroussilhe, Gesmundo, Attariyan, and Gelly]{houlsby2019parameter}
N.~Houlsby, A.~Giurgiu, S.~Jastrzebski, B.~Morrone, Q.~De~Laroussilhe, A.~Gesmundo, M.~Attariyan, and S.~Gelly, ``Parameter-efficient transfer learning for nlp,'' in \emph{International conference on machine learning}, 2019, pp. 2790--2799.

\bibitem[Hu et~al.(2021)Hu, Shen, Wallis, Allen-Zhu, Li, Wang, Wang, and Chen]{hu2021lora}
E.~J. Hu, Y.~Shen, P.~Wallis, Z.~Allen-Zhu, Y.~Li, S.~Wang, L.~Wang, and W.~Chen, ``Lora: Low-rank adaptation of large language models,'' \emph{arXiv preprint arXiv:2106.09685}, 2021.

\bibitem[Zhou et~al.(2024)Zhou, Sun, Ning, Ye, and Zhan]{zhou2024continual}
D.-W. Zhou, H.-L. Sun, J.~Ning, H.-J. Ye, and D.-C. Zhan, ``Continual learning with pre-trained models: A survey,'' \emph{arXiv preprint arXiv:2401.16386}, 2024.

\bibitem[Wang et~al.(2022{\natexlab{a}})Wang, Zhang, Lee, Zhang, Sun, Ren, Su, Perot, Dy, and Pfister]{wang2022learning}
Z.~Wang, Z.~Zhang, C.-Y. Lee, H.~Zhang, R.~Sun, X.~Ren, G.~Su, V.~Perot, J.~Dy, and T.~Pfister, ``Learning to prompt for continual learning,'' in \emph{Proceedings of the IEEE/CVF conference on computer vision and pattern recognition}, 2022, pp. 139--149.

\bibitem[Wang et~al.(2022{\natexlab{b}})Wang, Zhang, Ebrahimi, Sun, Zhang, Lee, Ren, Su, Perot, Dy, et~al.]{wang2022dualprompt}
Z.~Wang, Z.~Zhang, S.~Ebrahimi, R.~Sun, H.~Zhang, C.-Y. Lee, X.~Ren, G.~Su, V.~Perot, J.~Dy \emph{et~al.}, ``Dualprompt: Complementary prompting for rehearsal-free continual learning,'' in \emph{European Conference on Computer Vision}, 2022, pp. 631--648.

\bibitem[Wang et~al.(2022{\natexlab{c}})Wang, Huang, and Hong]{wang2022s-prompt}
Y.~Wang, Z.~Huang, and X.~Hong, ``S-prompts learning with pre-trained transformers: An occam’s razor for domain incremental learning,'' \emph{Advances in Neural Information Processing Systems}, vol.~35, pp. 5682--5695, 2022.

\bibitem[Jung et~al.(2023)Jung, Han, Bang, and Song]{jung2023generating}
D.~Jung, D.~Han, J.~Bang, and H.~Song, ``Generating instance-level prompts for rehearsal-free continual learning,'' in \emph{Proceedings of the IEEE/CVF International Conference on Computer Vision}, 2023, pp. 11\,847--11\,857.

\bibitem[Fine and Scheinberg(2001)]{fine2001efficient}
S.~Fine and K.~Scheinberg, ``Efficient svm training using low-rank kernel representations,'' \emph{Journal of Machine Learning Research}, vol.~2, no. Dec, pp. 243--264, 2001.

\bibitem[Li et~al.(2023{\natexlab{b}})Li, Li, Savarese, and Hoi]{li2023blip}
J.~Li, D.~Li, S.~Savarese, and S.~Hoi, ``Blip-2: Bootstrapping language-image pre-training with frozen image encoders and large language models,'' in \emph{International conference on machine learning}, 2023, pp. 19\,730--19\,742.

\bibitem[Maddern et~al.(2017)Maddern, Pascoe, Linegar, and Newman]{maddern20171}
W.~Maddern, G.~Pascoe, C.~Linegar, and P.~Newman, ``1 year, 1000 km: The oxford robotcar dataset,'' \emph{The International Journal of Robotics Research}, vol.~36, no.~1, pp. 3--15, 2017.

\bibitem[Kim et~al.(2020)Kim, Park, Cho, Jeong, and Kim]{kim2020mulran}
G.~Kim, Y.~S. Park, Y.~Cho, J.~Jeong, and A.~Kim, ``Mulran: Multimodal range dataset for urban place recognition,'' in \emph{2020 IEEE international conference on robotics and automation (ICRA)}, 2020, pp. 6246--6253.

\bibitem[Zhou et~al.(2022)Zhou, Liu, Qiao, Xiang, and Loy]{zhou2022domain}
K.~Zhou, Z.~Liu, Y.~Qiao, T.~Xiang, and C.~C. Loy, ``Domain generalization: A survey,'' \emph{IEEE transactions on pattern analysis and machine intelligence}, vol.~45, no.~4, pp. 4396--4415, 2022.

\bibitem[Liu et~al.(2021)Liu, Zhang, Hou, Mian, Wang, Zhang, and Tang]{liu2021self}
X.~Liu, F.~Zhang, Z.~Hou, L.~Mian, Z.~Wang, J.~Zhang, and J.~Tang, ``Self-supervised learning: Generative or contrastive,'' \emph{IEEE transactions on knowledge and data engineering}, vol.~35, no.~1, pp. 857--876, 2021.

\bibitem[Aljundi et~al.(2019)Aljundi, Lin, Goujaud, and Bengio]{aljundi2019gradient}
R.~Aljundi, M.~Lin, B.~Goujaud, and Y.~Bengio, ``Gradient based sample selection for online continual learning,'' \emph{Advances in neural information processing systems}, vol.~32, 2019.

\end{thebibliography}
\bibliographystyle{IEEEtranN}

\section*{Biography Section}
\begin{IEEEbiography}[{\includegraphics[width=1in,height=1.25in,clip,keepaspectratio]{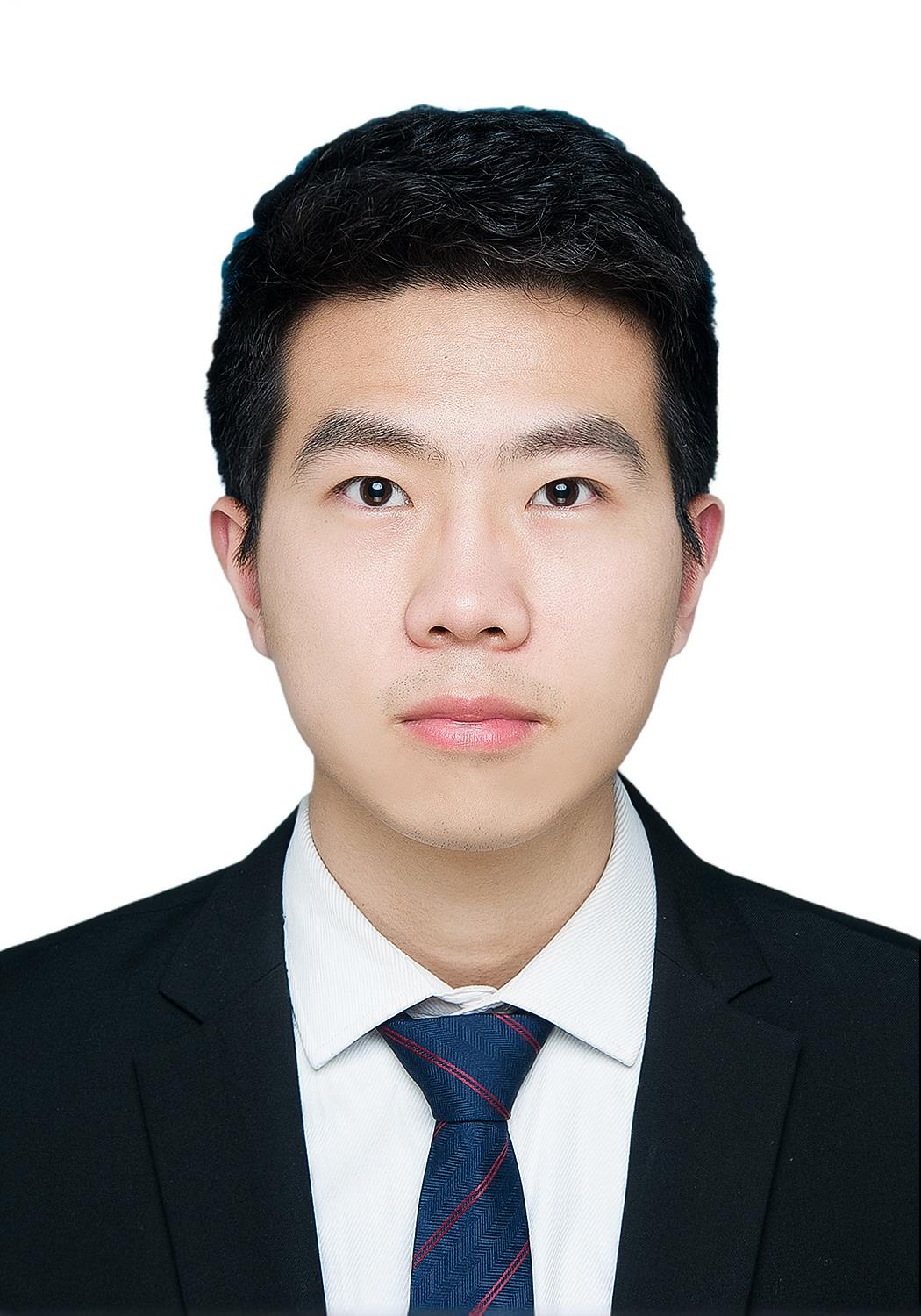}}]{Xianghong Zou}
received the B.S. degree in geomatics engineering, the M.S. degree, and the Ph.D. degree in photogrammetry and remote sensing from Wuhan University, Wuhan, China, in 2016, 2019, and 2024, respectively.

He is currently a post-doctoral researcher with the School of Advanced Manufacturing, Nanchang University, Nanchang, China. His research interests include point cloud data processing, global localization, and 3D change detection.
\end{IEEEbiography}

\begin{IEEEbiography}[{\includegraphics[width=1in,height=1.25in,clip,keepaspectratio]{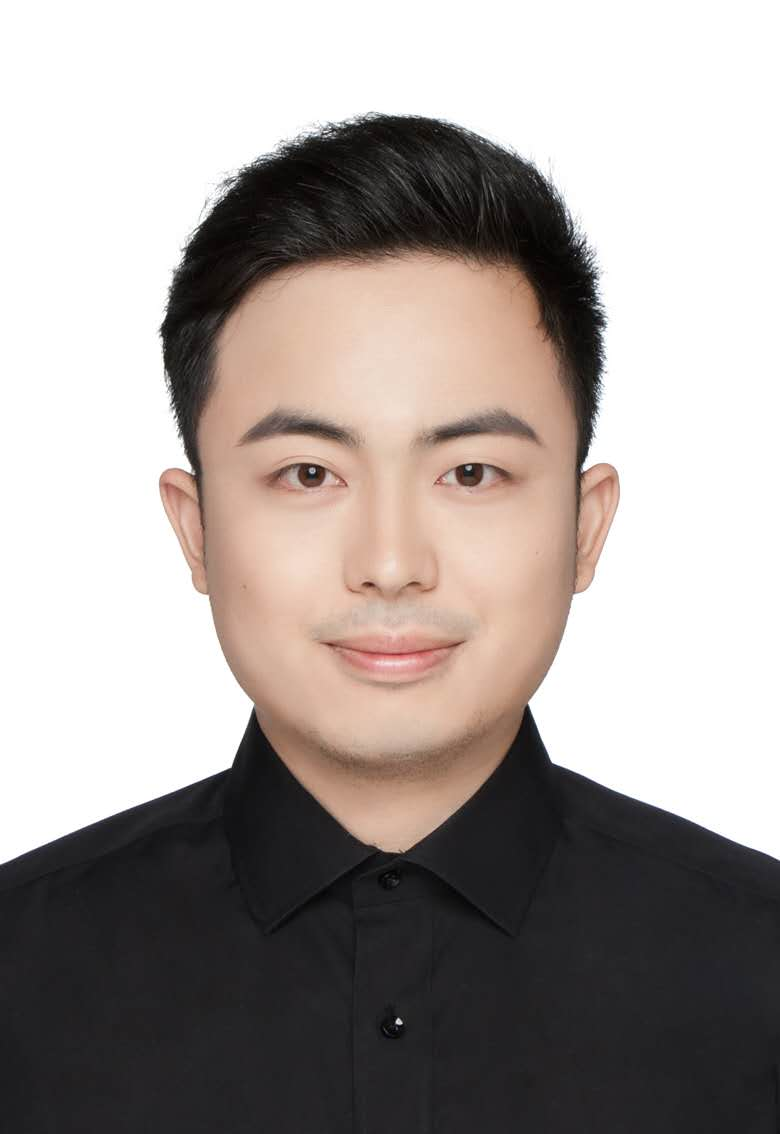}}]{Jianping Li}
received the B.S. degree in GIS and the Ph.D. degree in photogrammetry and remote sensing from Wuhan University, Wuhan, China, in 2015 and 2021, respectively. He is currently a Research Fellow with the School of Electrical and Electronic Engineering, Nanyang Technological University, Singapore. 

His research interests include 3D sensing system integration, UAV/UGV mapping, robot perception, and point cloud data processing.
\end{IEEEbiography}

\begin{IEEEbiography}[{\includegraphics[width=1in,height=1.25in,clip,keepaspectratio]{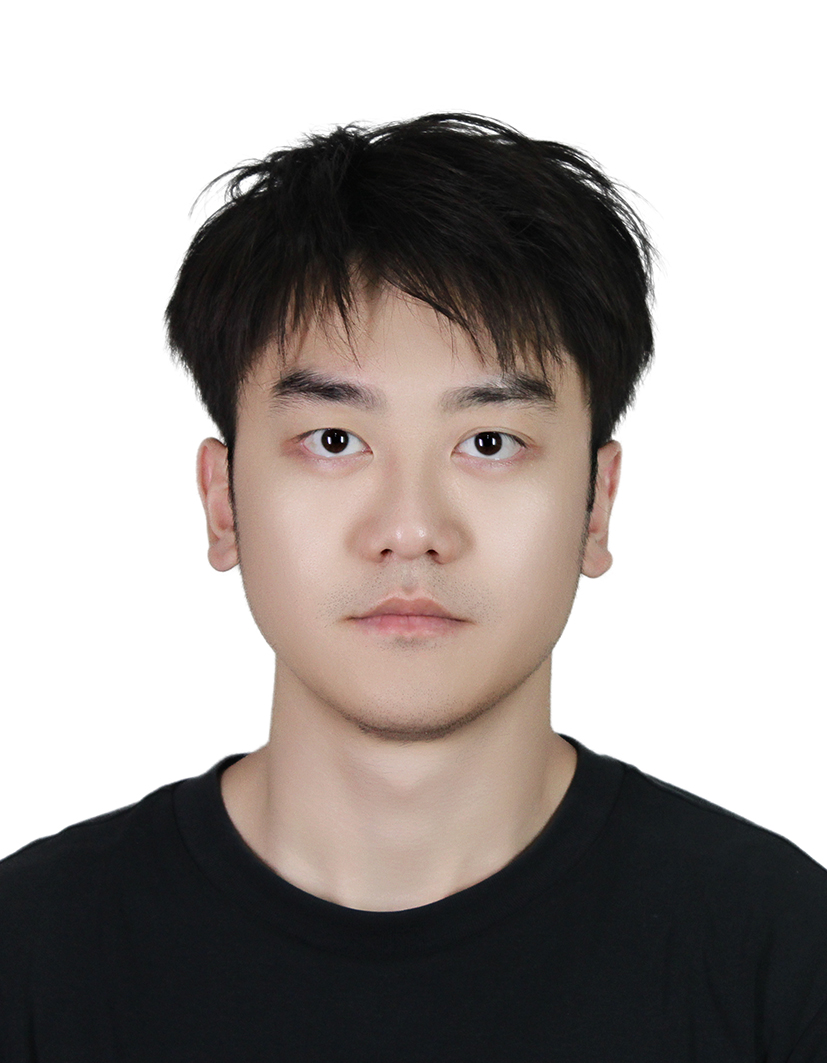}}]{Zhe Chen}
received the B.S. degree from the School of Earth Sciences and Engineering, Hohai University, Nanjing, China, in 2020. He is currently pursuing the Ph.D. degree with the State Key Laboratory of Information Engineering in Surveying, Mapping, and Remote Sensing, Wuhan University, Wuhan, China.

His research interests include image/point cloud processing and their applications in urban morphology.
\end{IEEEbiography}

\begin{IEEEbiography}[{\includegraphics[width=1in,height=1.25in,clip,keepaspectratio]{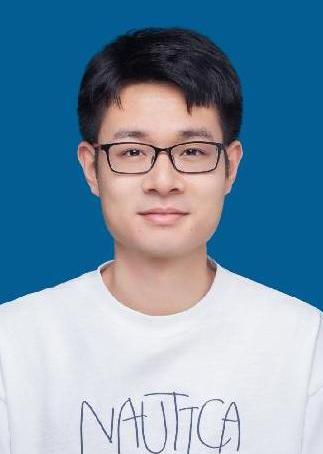}}]{Zhen Cao}
is a first-year Ph.D. student at the State Key Laboratory of Information Engineering in Surveying, Mapping and Remote Sensing (LIESMARS), Wuhan University, under the supervision of Prof. Zhen Dong and Prof. Bisheng Yang. He obtained his B.Eng. degree in Photogrammetry and Remote Sensing from the School of Geodesy and Geomatics (SGG), Wuhan University.
His research interest lies in the field of 3D Computer Vision, particularly including point cloud completion, scene understanding, and intelligent transportation systems (ITS).
\end{IEEEbiography}

\begin{IEEEbiography}[{\includegraphics[width=1in,height=1.25in,clip,keepaspectratio]{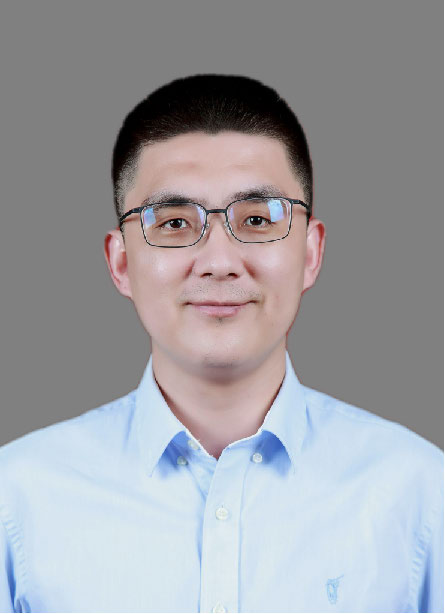}}]{Zhen Dong}
received the B.E. and Ph.D. degrees in remote sensing and photogrammetry from Wuhan University, in 2011 and 2018, respectively. He is currently a Professor of the State Key Laboratory of Information Engineering in Surveying, Mapping and Remote Sensing (LIESMARS), Wuhan University.

His research interest lies in the field of 3D Computer Vision, particularly including 3D reconstruction, scene understanding, point cloud processing as well as their applications in the intelligent transportation systems, digital twin cities, urban sustainable development and robotics.
\end{IEEEbiography}

\begin{IEEEbiography}[{\includegraphics[width=1in,height=1.25in,clip,keepaspectratio]{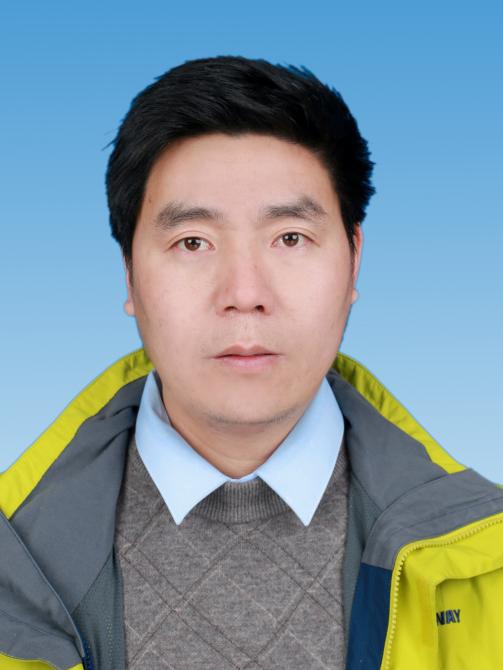}}]{Qiegen Liu}
received the Ph.D. degree in biomedical engineering from Shanghai Jiaotong University, Shanghai, China, in 2012.

During his time, he has been a Guest Speaker with the Institute of Computational Biology, Chinese Academy of Sciences (CAS), Beijing, China, and the Laubert Center for Medical Imaging, Shenzhen Institute of Advanced Technology (SIAT), Shenzhen, China. He did his Postdoctoral Work with UIUC, Champaign, IL, USA, in 2015, and the University of Calgary, Calgary, AB, Canada, in 2016. He is mainly engaged in sparse and deep learning representations and their applications in medical imaging and image processing.
\end{IEEEbiography}

\begin{IEEEbiography}[{\includegraphics[width=1in,height=1.25in,clip,keepaspectratio]{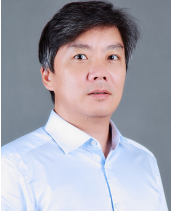}}]{Bisheng Yang}
received the B.S. degree in engineering survey, the M.S. degree, and the Ph.D. degree in photogrammetry and remote sensing from Wuhan University, China, in 1996, 1999, and 2002, respectively. From 2002 to 2006, he held a post-doctoral position at the University of Zurich, Switzerland. Since 2007, he has been a Professor with the State Key Laboratory of Information Engineering in Surveying, Mapping and Remote Sensing (LIESMARS), Wuhan University, where he is currently the Director of LIESMARS.

His main research interests comprise 3-D geographic information systems, urban modeling, and digital city. He was a Guest Editor of the ISPRS Journal of Photogrammetry and Remote Sensing, and Computers \& Geosciences.
\end{IEEEbiography}


\end{document}